\documentclass[]{fairmeta}

\usepackage{amsmath,amsfonts,bm}

\def\eqref#1{equation~\ref{#1}}

\def\1{\bm{1}}

\DeclareMathAlphabet{\mathsfit}{\encodingdefault}{\sfdefault}{m}{sl}
\SetMathAlphabet{\mathsfit}{bold}{\encodingdefault}{\sfdefault}{bx}{n}

\usepackage{url}
\usepackage{xcolor}
\usepackage{colortbl}
\usepackage{array}
\usepackage{afterpage}
\usepackage{capt-of}
\usepackage[export]{adjustbox}
\usepackage{tikz}
\usetikzlibrary{arrows.meta,positioning,fit,backgrounds,calc,decorations.pathreplacing}
\microtypesetup{expansion=false}
\let\metaincludegraphics\includegraphics
\renewcommand{\includegraphics}[2][]{%
  \metaincludegraphics[#1,max height=0.78\textheight,keepaspectratio]{#2}}

\title{FLAT: Resampling Image and Text into 1D \textbf{F}lexible-\textbf{L}ength \textbf{A}ligned \textbf{T}ransmodal Tokens for Retrieval and Generation}

\author[1]{Guangyu Sun}
\author[1]{Shlok Kumar Mishra}
\author[1]{Wentao Bao}
\author[1]{Robert Zhenheng Yang}
\author[1]{Xiao Wang}
\author[1]{Xiyuan Wang}
\author[1]{Yujunrong Ma}
\author[1]{Chen Yuan}
\author[1]{Max Xiangjun Fan} 
\author[1]{Jun Xiao}
\author[1]{Jianpeng Cheng}
\affiliation[]{Meta AI}

\metadata[Project Website]{\url{https://guangyusun.com/flat-website}}
\correspondence{\email{guangyu@meta.com} \email{jianpengcheng@meta.com}}

\abstract{Traditional multimodal representation learning and generation are two stages: a contrastive or self-supervised visual encoder is trained first, followed by a separate downstream generative model. This setup bottlenecks generative performance behind frozen embeddings. To bridge this gap, we revisit joint multimodal representation learning and generation to produce linearly interpolatable embeddings that are directly consumable by generative decoders. We present FLAT (\textbf{F}lexible-\textbf{L}ength \textbf{A}ligned \textbf{T}ransmodal representations), a representation pre-training framework that jointly optimizes a shared multimodal encoder alongside downstream text-to-image (T2I) and image-to-text (I2T) decoders. By combining contrastive alignment with bidirectional cross-modal generative objectives, FLAT ensures its representations function  as both discriminative semantic descriptors and generative conditions. Architecturally, FLAT maps visual and textual inputs into a unified continuous 1D sequence space, applying nested dropout over prefix-$K$ tokens to enable dynamic output lengths. A single pre-training stage allows FLAT to perform cross-modal retrieval and generation across variable prefix $K$, achieving a T2I GenEval score of 71.1. Task-specific fine-tuning aligns model performance with state-of-the-art baselines: 83.1 GenEval on T2I generation; 40.5 BLEU-4 and 138.6 CIDEr on MS-COCO image captioning; and Recall@5 scores of 86.8 (I2T) / 75.8 (T2I) on MS-COCO alongside 98.3 (I2T) / 93.6 (T2I) on Flickr30K. Finally, qualitative evaluations demonstrate that FLAT representations natively support linear interpolation, latent space arithmetic, and zero-shot composed retrieval.}

\date{September 2026}

\begin{document}

\maketitle

\section{Introduction}
\label{sec:intro}

Traditional multimodal architectures  separate modality representation learning from cross-modal generation. In representation learning, models like CLIP~\citep{radford2021clip} align visual and textual features using contrastive objectives, or DINO~\citep{caron2021emerging} and JEPA~\citep{assran2023self} train visual with self-supervision. In image-to-text (I2T) models, the frozen visual representations are connected to a transformer decoder and re-aligned with the text embedding space~\citep{liu2023llava}. Conversely, text-to-image (T2I) models rely on pre-trained, frozen text encoders: Stable Diffusion and SDXL utilize CLIP text embeddings, SD3 combines CLIP and T5,  PixArt and Sana and MetaQuery leverage frozen large language models~\citep{radford2021clip,rombach2022high,podell2024sdxl,esser2024sd3,chen2024pixart,xie2025sana,pan2025metaquery}. This decoupling of representation learning and generative modeling bottlenecks generative performance behind frozen representations and requires re-alignment in generative model training.

To bridge this gap, recent unified multimodal models either jointly train representation encoders with transformer backbones using generative objectives, or bypass vision encoders to operate directly in pixel space~\citep{agrawal2024pixtral,diao2025evev2,liu2026tuna}. In this work, we revisit an alternative direction: retaining an explicit contrastive, linearly-interpolatable embedding  space while simultaneously optimizing alignment and generation. We present \textbf{FLAT} (\textbf{F}lexible \textbf{L}ength \textbf{A}ligned \textbf{T}ransmodal representations), a representation pre-training framework that jointly trains a multimodal encoder with T2I and I2T decoders using \textbf{dual bidirectional alignment and generative} losses. Our central insight is that retrieval and generation are mutually reinforcing~\citep{yu2022coca}—effective representations should serve as both discriminative semantic descriptors and  generative conditions for either direction. We demonstrate that these combined objectives produce  representations that are linearly-interpolatable and consistently enhance downstream tasks.

FLAT builds upon recent advances in 1D visual tokenization, which resample image grids into compact 1D token sequences to eliminate spatial redundancy~\citep{yu2024titok,bachmann2025flextok}. Similar techniques have been employed to compress lengthy text into summary vectors~\citep{chevalier2023autocompressor}. 
Unlike prior visual tokenizers that focus on image reconstruction, FLAT resamples both visual and textual inputs into a unified 1D representation space via a shared multimodal encoder, producing continuous 1D representations that are contrastively aligned during generative training. FLAT incorporates nested dropout~\citep{bachmann2025flextok,rippel2014nested,kusupati2022mrl} into sequence representation learning by optimizing over random prefix-K tokens per batch, enabling flexible sequence length selection at inference time.

We evaluate FLAT across T2I and I2T generation and retrieval. By varying the number of prefix-$K$ tokens, one pre-trained FLAT model retrieves and synthesizes coarse-to-fine cross-modal outputs from the same encoder pass. After a single pre-training stage it reaches a GenEval score of $71.1$ on T2I generation and, zero-shot, MS-COCO Recall@5 of $69.1$ (I2T) and $64.6$ (T2I). Task-specific fine-tuning then brings model performance on each task in line with published SOTA baselines: $83.1$ GenEval on T2I generation, above every baseline we compare against including two 7B models that additionally rewrite the prompt; $40.5$ BLEU-4 and $138.6$ CIDEr on MS-COCO for I2T generation; and Recall@5 of $86.8$ (I2T) / $75.8$ (T2I) on MS-COCO and $98.3$ (I2T) / $93.6$ (T2I) on Flickr30K. On ImageNet linear probing the frozen representation reaches $81.8$, above all published results for generative latent space. Finally, qualitative evaluations show that FLAT representations natively support interpolation, semantic arithmetic on latent space, and zero-shot composed retrieval.

\section{Related Work}
\label{sec:related}

\textbf{Decoupled Representation Learning and Generation.} Contrastive representation learning methods in the CLIP family~\citep{radford2021clip,jia2021scaling,zhai2023sigmoid,sun2023eva} have established the benchmark for cross-modal alignment and retrieval. However, downstream multimodal generative models typically build on separately pretrained encoders. In I2T architectures such as BLIP-2~\citep{li2023blip2} and LLaVA~\citep{liu2023llava}, frozen visual encoders are connected to large language models through trainable adapters. Conversely, T2I models such as unCLIP~\citep{ramesh2022unclip}, SDXL~\citep{podell2024sdxl}, SD3~\citep{esser2024sd3}, PixArt-$\alpha$~\citep{chen2024pixart}, and Sana~\citep{xie2025sana} use frozen text or multimodal encoders as conditioning modules.
This separation persists in many unified multimodal systems, which keep pretrained visual encoders, tokenizers, or VAEs frozen while training the components for understanding and generation~\citep{wu2024janus,team2024chameleon,xie2024show,zhou2025transfusion,pan2025metaquery,chen2025blip3}.

\textbf{Joint Representation Learning and Generation.} BLIP~\citep{li2022blip} and CoCa~\citep{yu2022coca} combine contrastive image--text alignment with text generation. MAGE~\citep{li2023mage} combines masked image generation with self-supervised representation learning, while DREAM~\citep{li2026dream} jointly optimizes image--text contrastive alignment and masked image generation.
Meanwhile, recent unified multimodal models process understanding and generation within a shared architecture using generative training~\citep{huang2025ming,liu2026tuna,diao2026sensenova}. Another line develops unified visual tokenizers or encoders that support both semantic understanding and visual reconstruction~\citep{wu2024vilau,ma2025unitok,qu2025tokenflow,zhang2025qlip,lin2025toklip,fan2025prism,yue2026uniflow,zhang2026openvision}.
FLAT instead jointly optimizes a variable-length continuous representation for contrastive retrieval and bidirectional cross-modal generation, covering both I2T and T2I.

\textbf{Resampled 1D Representations.} TiTok~\citep{yu2024titok}, FlexTok~\citep{bachmann2025flextok} and GigaTok~\citep{xiong2025gigatok} resample spatial image grids into compact 1D token sequences.  AutoCompressor~\citep{chevalier2023autocompressor} resamples long text inputs into summary vectors. BLIP-2~\citep{li2023blip} resamples image encoder outputs with learnable queries for image-text alignment.
MiniGPT-5~\citep{zheng2023minigpt}, DreamLLM~\citep{dong2024dreamllm}, and MetaQuery~\citep{pan2025metaquery} map or query multimodal LLM hidden states to condition image generation.
In comparison, FLAT resamples both images and text into a shared continuous space as  1D sequence representations.

\section{Method}
\label{sec:method}

\begin{figure}[t]
\centering
\includegraphics[width=\textwidth]{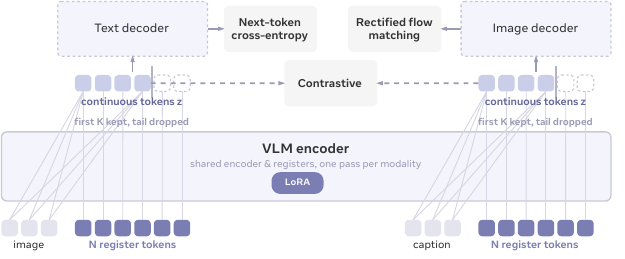}
\caption{\textbf{FLAT architecture overview.} A  shared VLM encoder processes visual and textual inputs through  register tokens, applying nested dropout to truncate sequences to variable prefix lengths during joint training. FLAT's unified 1D continuous representation simultaneously drives contrastive alignment, autoregressive caption generation, and rectified-flow image synthesis. }
\label{fig:arch}
\end{figure}

\subsection{Overview}
\label{sec:method:overview}

FLAT is a representation pre-training framework which learns a multimodal encoder and two transmodal decoders jointly with contrastive and generative objectives. Given an input text or image, FLAT maps the input to an ordered sequence of continuous tokens, which we call the \emph{representation}. The representation is contrastively aligned and read by both transmodal decoders. An overview of FLAT architecture is shown in Fig.~\ref{fig:arch}.

\textbf{Representation Encoder.}
let $x$ denote an input of modality $m \in \{\mathrm{img},\mathrm{txt}\}$. FLAT appends $N$ learnable register tokens $R = \{r_n\}_{n=1}^{N}$ to each input and encode the resulting sequence with a shared multimodal encoder $f_{\mathrm{enc}}$, which is a pre-trained Vision-Language Model (VLM) with a system prompt \textit{Represent the input}. The hidden states at the register positions are linearly projected to obtain the representation $\mathbf{z}$:
\begin{equation}
    \mathbf{z} = W_{\mathrm{lat}}\, f_{\mathrm{enc}}(x, R) \in \mathbb{R}^{N \times d},
    \label{eq:flat_encoding}
\end{equation}
where $d$ is the register dimension and $W_{\mathrm{lat}}$ the projection. The two modalities share the same registers and the same projection.

\textbf{Representation Dropout.}
To enable variable sequence lengths, we apply nested dropout~\citep{rippel2014nested} directly over the representation $\mathbf{z}$. For each training batch, a keep-length $K$ is sampled from a geometric ladder $\mathcal{K} = \{1, 4, 16, 64, 256\}$, and only the prefix sequence $\mathbf{z}_{:K}$ is passed to the contrastive objective and downstream decoders. Sampling from a geometric ladder rather than continuous integers avoids wasting optimization steps on imperceptible length differences. By default, FLAT samples uniformly from $\mathcal{K}$. Appendix~\ref{sec:app:k-sampling} compares uniform sampling with other sampling strategies.

Compared to FlexTok~\citep{bachmann2025flextok}, which maintains a fixed sequence length by padding tail positions with learned null embeddings, we find that directly truncating the representation to the prefix $\mathbf{z}_{:K}$ yields better results in FLAT training setups (Appendix~\ref{app:padding}). To ensure consistency, the target length $K$ is sampled once per training step and broadcast globally across all ranks. This guarantees that both contrastive and generative objectives operate on the exact same sequence prefix at every optimization step.

\textbf{Decoders.}
The truncated representation sequence $\mathbf{z}_{:K}$ serves as the semantic condition for both the I2T and T2I decoders. Given a paired image-text input, the visual representation $\mathbf{z}_{:K}^{\mathrm{img}}$ conditions caption generation, while the textual representation $\mathbf{z}_{:K}^{\mathrm{txt}}$ conditions image synthesis. For each generative pathway, $\mathbf{z}_{:K}$ is treated as a sequence of soft tokens and mapped into the respective decoder's input dimension $h$:
\begin{equation}
\mathbf{e} = \mathrm{RMSNorm}(\mathrm{MLP}(\mathbf{z}_{:K})) ;\in; \mathbb{R}^{K \times h},
\label{eq:sem}
\end{equation}
where the $\mathrm{MLP}$ projections consist of decoupled, task-specific parameters for the I2T and T2I decoders.

The I2T decoder is a standard autoregressive language model. The visual soft tokens $\mathbf{e}^{\mathrm{img}}$ are prepended with a standard system prompt (\textit{"Describe the input"}) and fed as input tokens to autoregressively generate the corresponding text caption.
The T2I decoder is a standard rectified flow transformer that denoises continuous VAE latents. The textual soft tokens $\mathbf{e}^{\mathrm{txt}}$ act as semantic conditioning signals integrated via cross-attention mechanisms, guiding the flow-matching model to synthesize VAE latents decodable into the target image.
The detailed, multi-task training objectives governing these decoders are formally defined in the next section.

\afterpage{
\begin{figure}[t]
\centering
\begin{minipage}[t]{0.49\textwidth}
\vspace{0pt}
\centering
\captionof{table}{\textbf{Text-to-image synthesis on GenEval.} FLAT achieves competitive scores while enabling inference-time sequence length selection. Evaluated from a single fine-tuned checkpoint across prefix lengths. $\dagger$ marks a system that rewrites the prompt before generating.}
\label{tab:res:geneval}
\scriptsize
\setlength{\tabcolsep}{3pt}
\begin{tabular}{@{}llc@{}}
\toprule
Method & Backbone & GenEval \\
\midrule
Chameleon~\citep{team2024chameleon}          & scratch 7B    & 0.39 \\
Janus~\citep{wu2024janus}                    & DeepSeek 1.5B & 0.61 \\
Transfusion~\citep{zhou2025transfusion}      & scratch 7B    & 0.63 \\
JanusFlow~\citep{ma2024janusflow}            & DeepSeek 1.5B & 0.63 \\
Emu3$^{\dagger}$~\citep{wang2024emu3}        & scratch 7B    & 0.66 \\
Show-o-512~\citep{xie2024show}               & Phi-1.5 1.3B  & 0.68 \\
Janus-Pro-1B~\citep{chen2025januspro}        & DeepSeek 1.5B & 0.73 \\
MetaQuery-L~\citep{pan2025metaquery} & Qwen2.5-VL 3B & 0.78 \\
Janus-Pro-7B~\citep{chen2025januspro}        & DeepSeek 7B   & 0.80 \\
MetaQuery-XL$^{\dagger}$~\citep{pan2025metaquery} & Qwen2.5-VL 7B & 0.80 \\
\midrule
FLAT, $K{=}1$    & Qwen3.5 2B & 0.49 \\
FLAT, $K{=}4$    & Qwen3.5 2B & 0.77 \\
FLAT, $K{=}16$   & Qwen3.5 2B & \textbf{0.83} \\
FLAT, $K{=}64$   & Qwen3.5 2B & \textbf{0.83} \\
FLAT, $K{=}256$  & Qwen3.5 2B & 0.82 \\
\bottomrule
\end{tabular}
\end{minipage}\hfill
\begin{minipage}[t]{0.48\textwidth}
\vspace{0pt}
\centering
\includegraphics{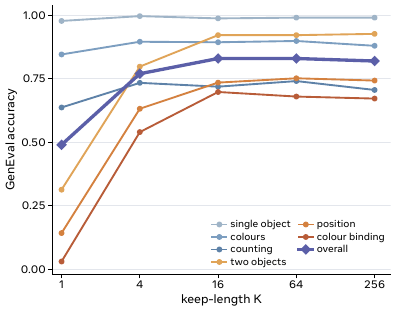}
\captionof{figure}{\textbf{GenEval accuracy across keep-lengths broken down by eval category.} Smaller $K$ can capture core entity and color, but larger $K$ is needed to capture spatial relations, multi-object layouts and attribute bindings.}
\label{fig:res:geneval:cat}
\end{minipage}

\vspace{8pt}

\includegraphics[width=\textwidth]{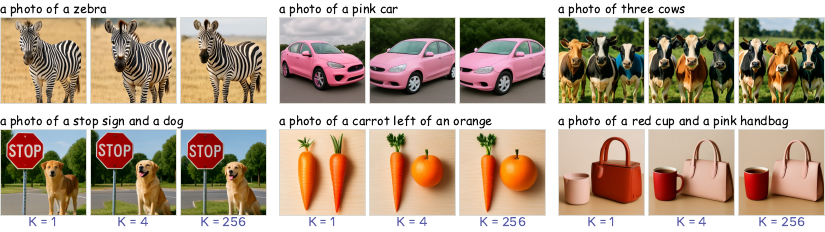}
\captionof{figure}{\textbf{Qualitative T2I synthesis across different prefix lengths.} Large $K$ provides more fine-grained visual details.}
\label{fig:res:geneval:grid}
\end{figure}

\begin{figure}[t]
\centering
\captionof{table}{\textbf{Image captioning on the MS-COCO Karpathy test split.} FLAT matches or outperforms competitive baselines with less latent features. Column \emph{Latent Feat.} represents latent feature shape which the text decoder is conditioned on. The metrics follow previous work: B@4 (BLEU-4), M (METEOR), R (ROUGE-L), C (CIDEr), S (SPICE). }
\label{tab:res:caption}
\small
\setlength{\tabcolsep}{6pt}
\begin{tabular}{lrrrrrr}
\toprule
Method & Latent Feat. & B@4 & M & R & C & S \\
\cmidrule(r){1-1}\cmidrule(lr){2-2}\cmidrule(l){3-7}
CrossFlow~\citep{liu2025crossflow}     & $4{\times}32{\times}32$ & 36.4 & 27.8 & 57.1 & 116.2 & 20.4 \\
SCD-Net~\citep{luo2023scdnet}          & $N{\times}512$    & 37.3 & 28.1 & 58.0 & 118.0 & 21.6 \\
Florence-2-L~\citep{xiao2024florence2} & --                & --   & --   & --   & 143.3 & --   \\
CoCa~\citep{yu2022coca}                & $256{\times}1408$ & 40.9 & 33.9 & --   & 143.6 & 24.7 \\
BLIP-2~\citep{li2023blip2}             & $32{\times}768$   & 43.7 & --   & --   & 145.8 & --   \\
\addlinespace[1.5pt]
\cmidrule(r){1-1}\cmidrule(lr){2-2}\cmidrule(l){3-7}
FLAT, $K{=}1$    & $1{\times}64$   & 35.3 & 28.7 & 57.7 & 122.0 & 22.0 \\
FLAT, $K{=}4$    & $4{\times}64$   & 38.1 & 29.9 & 59.4 & 131.0 & 23.3 \\
FLAT, $K{=}16$   & $16{\times}64$  & 39.6 & 30.8 & 60.4 & 136.2 & 24.1 \\
FLAT, $K{=}64$   & $64{\times}64$  & 40.2 & 31.1 & 60.8 & 138.1 & 24.4 \\
FLAT, $K{=}256$  & $256{\times}64$ & \textbf{40.5} & \textbf{31.2} & \textbf{60.8} & \textbf{138.6} & \textbf{24.4} \\
\bottomrule
\end{tabular}

\vspace{10pt}

\centering
\includegraphics[width=\textwidth]{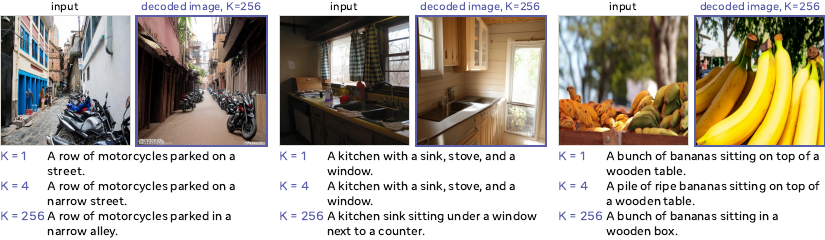}
\captionof{figure}{\textbf{Image captioning with fine-tuned I2T decoder.} Small $K$ generates semantically meaningful captions and larger $K$ refines lexical details. }
\label{fig:res:i2t}
\end{figure}
}

\subsection{Joint Training}
\label{sec:method:joint}
Given a paired input text-image, FLAT is trained with both bidirectional contrastive  loss and  generation loss. Since soft tokens $\mathbf{e}$ is projected from representation $\mathbf{z}$, we use $\mathbf{z}$ consistently as conditions in loss functions. 

\textbf{Contrastive Loss.} We contrast corresponding register positions in the representation $\mathbf{z}_{:K}$. For an image--text pair $(i,j)$ we define the similarity computed over their total
$K$ register positions as a late-interaction score between registers at matching positions:
\begin{equation}
\textstyle
    s_{ij}^{(K)} = \frac{1}{K}\sum_{n=1}^{K} \frac{ (z_{i,n}^{\mathrm{img}})^\top\hat z_{j,n}^{\mathrm{txt}} }{ \lVert z_{i,n}^{\mathrm{img}}\rVert_2\, \lVert z_{j,n}^{\mathrm{txt}}\rVert_2 },
    \label{eq:flat_sid_similarity}
\end{equation}
and the contrastive loss $\mathcal{L}_{\mathrm{align}}^{(K)}$ as
\begin{align}
\textstyle
    \mathcal{L}_{\mathrm{i2t}}^{(K)} = -\log\frac{\exp\left(s_{i,i}^{(K)}\right)}{\sum_{j=1}^{B_{\mathrm{glob}}}\exp\left(s_{i,j}^{(K)}\right)}, \quad &\mathcal{L}_{\mathrm{t2i}}^{(K)} = -\log\frac{\exp\left(s_{i,i}^{(K)}\right)}{\sum_{j=1}^{B_{\mathrm{glob}}}\exp\left(s_{j,i}^{(K)}\right)}, \label{eq:flat_sid_i2t_t2i}\\
    \mathcal{L}_{\mathrm{align}} &= \tfrac{1}{2}\left(\mathcal{L}_{\mathrm{i2t}}^{(K)}+\mathcal{L}_{\mathrm{t2i}}^{(K)}\right). \label{eq:flat_sid_alignment_loss}
\end{align}
where $B_{\mathrm{glob}}$ denotes the size of the gathered global candidates from all ranks. 

\textbf{Generation Loss.} For the same pair of image-text, we apply symmetrical generation losses for I2T and T2I respectively, both using the representations $\mathbf{z}_{:K}$ as semantic conditions. 

For I2T generation and an image input with representation $\mathbf{z}_{:K}^{\mathrm{img}}$, the loss over target caption $y$ and decoder parameter $\theta$ is defined as:
\begin{equation}
\textstyle
    \mathcal{L}_{\mathrm{txt}} = -\sum_t \log p_{\theta}\!\left( y_{t}\,\middle|\,y_{<t},\mathbf{z}_{:K}^{\mathrm{img}} \right)
    \label{eq:flat_sid_caption_loss}
\end{equation}

For T2I generation and a text input,  the representation $\mathbf{z}_{:K}^{\mathrm{txt}}$  are used as semantic conditions for a flow-matching decoder. Let $\hat x$ be the clean VAE image latent, $\epsilon\sim\mathcal{N}(0,I)$, and $t\in[0,1]$. The noised VAE latent and target velocity used in flow matching are defined as
\begin{equation}
    \hat x_t=(1-t)\hat x+t\epsilon, \qquad v^{\star}=\epsilon-\hat x.
    \label{eq:flat_sid_flow_path}
\end{equation}
The conditional flow-matching objective over decoder parameter $\phi$ is defined as:
\begin{equation}
    \mathcal{L}_{\mathrm{img}} = \mathbb{E}_{\hat x,\epsilon,t}\!\left[ \left\lVert v_{\phi}\!\left(\hat x_t,t,\mathbf{z}_{:K}^{\mathrm{txt}}\right) -(\epsilon-\hat x) \right\rVert_2^2 \right].
    \label{eq:flat_sid_image_loss}
\end{equation}

\textbf{Training Objective.}
In representation pre-training, FLAT  optimizes the total loss:
\begin{equation}
\mathcal{L}_{\mathrm{total}}=\lambda_{\mathrm{align}}\mathcal{L}_{\mathrm{align}} +\lambda_{\mathrm{txt}}\mathcal{L}_{\mathrm{txt}} +\lambda_{\mathrm{img}}\mathcal{L}_{\mathrm{img}}
\end{equation}
In  fine-tuning, we show that each task-specific objective can be optimized separately over one of the encoder and decoders to maximize model performance.

\section{Experiments}
\label{sec:exp}

We evaluate FLAT across generation, retrieval, and probing tasks to assess its unified 1D continuous representations. Our evaluation spans three core dimensions:
(1)~\textbf{Generative Evaluation,} demonstrating that the FLAT representation drives both T2I synthesis and I2T captioning competitively against specialized baselines;
(2)~\textbf{Discriminative Evaluation,} evaluating the FLAT representation directly with cross-modal retrieval, linear probing, and unsupervised clustering; and
(3)~\textbf{Geometry Analysis,} analyzing latent geometry to show that FLAT projects image and text into a unified representation space, natively unlocking emergent zero-shot linear interpolation,  latent space arithmetic and composed retrieval.
App.~\ref{app:crossdomain} further extends this evaluation to zero-shot generalization across multilingual prompts and video modalities.

\subsection{Implementation Details}
\label{sec:exp:impl}

The representation encoder and text decoder are built on a Qwen3.5-2B backbone~\citep{team2026qwen3} with different LoRA adapters, while the image decoder is initialized from SANA-1.6B~\citep{xie2025sana}. Full architectural, pre-training, and task-specific adaptation configurations are provided in App.~\ref{app:config}, together with a summary of the fine-tuning recipes in Table~\ref{tab:app:finetune}. All evaluations are conducted as a function of prefix length $K$. Specifically, we benchmark T2I synthesis on GenEval~\citep{ghosh2023geneval} without prompt rewriting; I2T captioning on the MS-COCO Karpathy test split; cross-modal retrieval on both the MS-COCO and Flickr30k Karpathy splits~\citep{karpathy2015deep}; and composed retrieval on the MMEB CIRR test split~\citep{jiang2024vlm2vecmmeb,liu2021cirr} (see App.~\ref{app:eval} for full eval protocols). While our main result focuses on the final fine-tuned checkpoints, we report pre-trained, zero-shot performance across all tasks in App.~\ref{app:zeroshot} and task-only adaptation without FLAT pre-training in App.~\ref{app:pretraining-transfer}.

\subsection{Generative Evaluation}
\label{sec:exp:gen}


\subsubsection{Text-to-Image Generation}
\label{sec:exp:t2i}

For T2I generation, we continue training the image decoder from the pre-trained checkpoint for $8$k steps on 120K clean image--text pairs (Table~\ref{tab:app:finetune}). As shown in Table~\ref{tab:res:geneval}, FLAT achieves a GenEval score of \textbf{0.83} using a 2B encoder without prompt rewriting, compared with 0.78 for MetaQuery-L under the same no-rewrite protocol. Categorical GenEval analysis further demonstrates a clear link between task complexity and required prefix length $K$ (Fig.~\ref{fig:res:geneval:cat}). While a single token ($K{=}1$) captures basic semantics like single object and color, relational and compositional tasks require larger prefixes. Complex categories---two objects, position, and color binding---score near zero at $K{=}1$ (e.g., 0.03 for color binding), but recover steeply by $K{=}4$ and $K{=}16$ (reaching 0.67). As shown in Fig.~\ref{fig:res:geneval:grid}, $K{=}1$ generates only the main entity, while spatial arrangements, secondary entities, and attribute-object bindings emerge accurately as $K$ increases.

\subsubsection{Image-to-Text Generation}
\label{sec:exp:i2t}

The I2T evaluation fine-tunes only the I2T-decoder LoRA with captioning loss on COCO Karpathy training+restval set for $5$k steps (Table~\ref{tab:app:finetune}). As detailed in Table~\ref{tab:res:caption}, performance improves steadily across all metrics as $K$ increases. Qualitatively (Fig.~\ref{fig:res:i2t}), a single token ($K{=}1$) generates well-formed, coherent captions, while additional tokens enhance lexical precision (e.g., street $\rightarrow$ alley) and concrete detail (e.g., a shelf $\rightarrow$ a shelf with baskets). These qualitative refinements directly mirror the quantitative gains in Table~\ref{tab:res:caption}. Even at $K{=}1$, FLAT outperforms baselines such as CrossFlow~\citep{liu2025crossflow} and SCD-Net~\citep{luo2023scdnet} across CIDEr, METEOR, and SPICE scores.




\subsection{Discriminative Evaluation}
\label{sec:exp:disc}

\subsubsection{Cross-Modal Retrieval}
\label{sec:exp:retrieval}

\begin{figure}[t]
\centering
\begin{minipage}[t]{0.44\textwidth}
\vspace{0pt}
\centering
\includegraphics{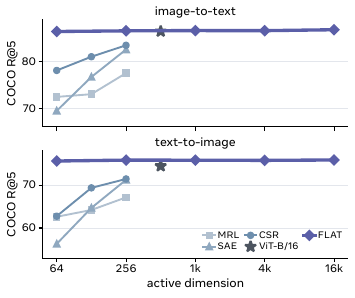}
\captionof{figure}{\textbf{MS-COCO retrieval accuracy across active dimensions.} Unlike other variable-width baselines whose performance degrades sharply at reduced dimensions, FLAT maintains stable retrieval accuracy across all dimensions.}
\label{fig:res:csr}
\end{minipage}\hfill
\begin{minipage}[t]{0.545\textwidth}
\vspace{0pt}
\centering
\captionof{table}{\textbf{T2I and I2T retrieval results across prefix $K$.} The first token captures key cross-modal discriminability, keeping retrieval accuracy virtually unaffected by prefix truncation. Performance is evaluated on the MS-COCO and Flickr30k Karpathy test splits.}
\label{tab:res:ret}
\small
\setlength{\tabcolsep}{5pt}
\begin{tabular}{lrrrrrr}
\toprule
& \multicolumn{3}{c}{I$\rightarrow$T} & \multicolumn{3}{c}{T$\rightarrow$I} \\
\cmidrule(lr){2-4}\cmidrule(l){5-7}
$K$ & R@1 & R@5 & R@10 & R@1 & R@5 & R@10 \\
\cmidrule(r){1-1}\cmidrule(lr){2-4}\cmidrule(l){5-7}
\multicolumn{7}{@{}l}{\emph{MS-COCO Karpathy 5k}}\\
1   & 63.00 & 86.44 & 92.50 & 47.76 & 75.59 & 84.67 \\
4   & 63.40 & 86.58 & 92.58 & 47.93 & 75.80 & 84.71 \\
16  & 63.30 & 86.64 & 92.56 & 47.94 & 75.78 & 84.72 \\
64  & 63.32 & 86.62 & 92.56 & 47.73 & 75.74 & 84.69 \\
256 & \textbf{63.54} & \textbf{86.82} & \textbf{92.62} & \textbf{47.98} & \textbf{75.83} & \textbf{84.78} \\
\addlinespace[1.5pt]
\cmidrule(r){1-1}\cmidrule(lr){2-4}\cmidrule(l){5-7}
\multicolumn{7}{@{}l}{\emph{Flickr30k Karpathy 1k}}\\
1   & 90.50 & 98.10 & 99.50 & 76.94 & 93.52 & 96.48 \\
4   & \textbf{91.20} & \textbf{98.30} & 99.50 & \textbf{77.40} & 93.58 & \textbf{96.54} \\
16  & 90.80 & 98.30 & \textbf{99.60} & 77.32 & 93.60 & 96.54 \\
64  & 90.90 & 98.20 & 99.40 & 76.88 & \textbf{93.72} & 96.54 \\
256 & 90.80 & 98.20 & 99.30 & 76.26 & 93.16 & 96.40 \\
\bottomrule
\end{tabular}
\end{minipage}
\vspace{-5mm}
\end{figure}

For each benchmark, retrieval adapts the encoder LoRA, registers, and latent projection using contrastive loss on the corresponding training split. The reported checkpoints use $5$k steps for COCO and $320$ steps for Flickr30K (Table~\ref{tab:app:finetune}). Fig.~\ref{fig:res:csr} compares FLAT against representations that also support adaptive dimensionality from~\citet{wen2025csr}, including MRL, SAE, and CSR. For FLAT, each token contains 64 dimensions, spanning from a 64-dimensional embedding ($K{=}1$) to a 16,384-dimensional embedding ($K{=}256$). Notably, FLAT's retrieval performance remains virtually invariant across varying $K$, whereas baseline accuracy degrades sharply as $K$ decreases. Table~\ref{tab:res:ret} reports detailed retrieval results across the $K$ spectrum. Remarkably, a single token is sufficient for retrieval: every metric stays within roughly one percentage point across a $256\times$ reduction in width. This mirrors our observations in T2I and I2T generation, confirming that the first token carries sufficient global semantics and discriminative power.

\subsubsection{Linear Probing and Clustering}
\label{sec:exp:probe}

\begin{figure}[t]
\centering
\begin{minipage}[t]{0.635\textwidth}
\vspace{0pt}
\centering
\includegraphics[width=\textwidth]{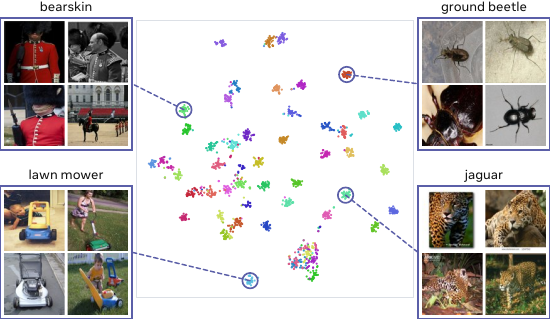}
\captionof{figure}{\textbf{Unsupervised $k$-means clustering with a single FLAT token.}
Clusters formed from frozen $K{=}1$ representations (64 dimensions) align with
ImageNet ground-truth classes without label supervision.}
\label{fig:res:cluster}
\end{minipage}\hfill
\begin{minipage}[t]{0.345\textwidth}
\vspace{0pt}
\centering
\captionof{table}{\textbf{ImageNet-1K linear probing top-1 accuracy.} FLAT outperforms both generative latent spaces and joint generative-contrastive models on ImageNet-1K linear probing. We evaluate this performance by fitting a single linear classifier on top of frozen, concatenated $K$ tokens.}
\label{tab:res:inet}
\scriptsize
\setlength{\tabcolsep}{3pt}
\begin{tabular}{@{}lrr@{}}
\toprule
Method & Dim & Top-1 \\
\cmidrule(r){1-1}\cmidrule(lr){2-2}\cmidrule(l){3-3}
\multicolumn{3}{@{}l}{\textit{Generative Latent Spaces}} \\
\quad SD-VAE latent                        & 4     & 8.0 \\
\quad REPA~\citep{leng2025repae}            & --    & 62.5 \\
\quad MAE-B~\citep{he2022mae}               & 768   & 68.0 \\
\addlinespace[2pt]
\multicolumn{3}{@{}l}{\textit{Jointly Contrastive and Generative}} \\
\quad DREAM~\citep{li2026dream}             & 1024  & 72.7 \\
\quad FLAT, $K{=}1$                         & 64    & 73.3 \\
\quad FLAT, $K{=}4$                         & 256   & 77.0 \\
\quad FLAT, $K{=}16$                        & 1024  & 81.2 \\
\quad FLAT, $K{=}64$                        & 4096  & \textbf{81.8} \\
\quad FLAT, $K{=}256$                       & 16384 & 81.8 \\
\bottomrule
\end{tabular}
\end{minipage}
\vspace{-5mm}
\end{figure}

\begin{figure}[t]
\centering
\includegraphics[width=\textwidth]{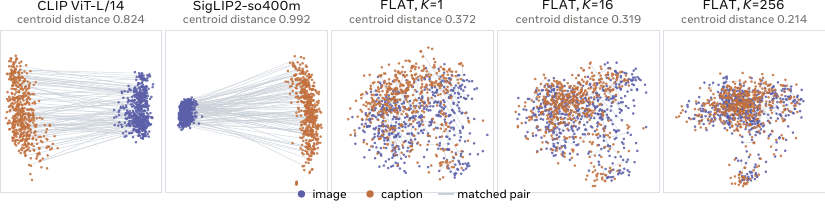}
\vspace{-3mm}
\caption{\textbf{PCA distributions for image-text pairs from MS-COCO.} Dual-encoder baselines exhibit clear spatial separation between image and text representations, whereas FLAT projects both modalities into a highly overlapping latent space. Increasing prefix $K$ further shrinks the cross-modal centroid distance.}
\label{fig:res:gap}
\vspace{-3mm}
\end{figure}

To directly evaluate the representation space, we perform linear probing and unsupervised clustering on FLAT representations. A single linear classifier trained on a frozen $K{=}1$ token achieves $73.3\%$ top-1 accuracy, scaling to $81.2\%$ at $K{=}16$. This outperforms all purely generative latent spaces and surpasses DREAM~\citep{li2026dream} which jointly trains the representation encoder with an image decoder, at equivalent dimensionality. Furthermore, without fitting any parametric heads, $k$-means clustering on a single FLAT token cleanly recovers clusters aligned with ImageNet classes (Fig.~\ref{fig:res:cluster}).

\subsection{Geometry Analysis}
\label{sec:exp:analysis}


\subsubsection{Geometry Visualization} 
 As shown in Fig.~\ref{fig:res:gap}, while dual-encoder models like CLIP and SigLIP2 suffer from a persistent modality gap despite strong retrieval performance~\citep{radford2021clip,tschannen2025siglip2,liang2022mindthegap}, FLAT projects both modalities into a tighter embedding space. FLAT achieves significantly greater cross-modal overlap: a single register token cuts CLIP's centroid distance by half, and the full 256 tokens reduces it to a quarter. This tightly aligned geometry enables zero-shot latent operations as below.


\subsubsection{Continuous Feature Interpolation}
\begin{figure}[t]
\centering
\includegraphics[width=\textwidth]{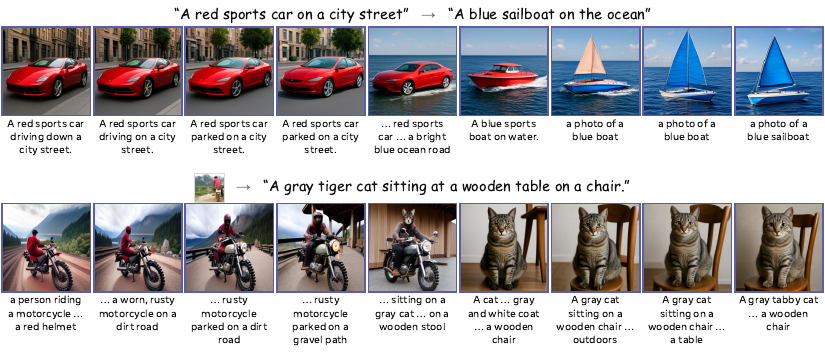}
\vspace{-5mm}
\caption{\textbf{Continuous cross-modal feature interpolations.} Linear interpolation between a pair of image--text produces smooth semantic transitions. Image and text decoders map the same  representation to corresponding concepts.}
\label{fig:res:interp}
\vspace{-3mm}
\end{figure}

We perform linear interpolation between a pair of FLAT codes as $z_{\alpha}=(1-\alpha)z_1+\alpha z_2$, and decode each intermediate latent representation using both generative heads. As shown in Fig.~\ref{fig:res:interp}, this yields smooth semantic transitions across the input image--text pairs. Both the image and text decoders generate intermediate concepts along the interpolation trajectory. Furthermore, the prefix length $K$ dictates transition granularity: longer prefixes smoothly vary fine-grained attributes, whereas $K{=}1$ forms a coarser semantic average (App.~\ref{app:interp}).

\subsubsection{Latent Space Arithmetic}
\begin{figure}[t]
\centering
\includegraphics[width=\textwidth]{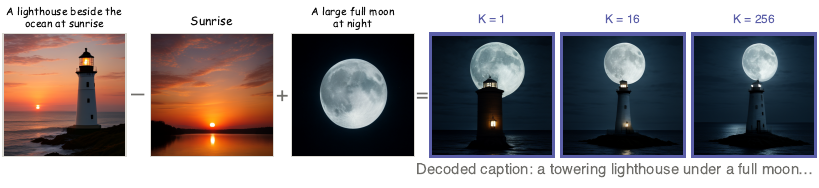}
\vspace{-5mm}
\caption{\textbf{Zero-shot token-wise arithmetic operation.} Linear addition and subtraction modify the target concept in both image and text outputs while preserving the surrounding scene.}
\label{fig:res:arith}
\vspace{-2mm}
\end{figure}

FLAT representations naturally support zero-shot  arithmetic, $z_{\text{edit}} = z_1 - z_2 + z_3$, without explicit editing supervision. As shown in Fig.~\ref{fig:res:arith}, subtracting \emph{sunrise} and adding \emph{a full moon at night} modifies the focal concept while preserving the surrounding scene context, with both decoders interpreting the composite representation consistently. Furthermore, these arithmetic operations can directly serve as queries for zero-shot composed image retrieval on CIRR (App.~\ref{app:cirr}).

\subsection{Ablation Study on Training Loss Combinations}
\label{sec:exp:ablation}

FLAT jointly optimizes a contrastive loss alongside bidirectional generation losses for I2T captioning and T2I synthesis. To evaluate the contribution of each objective, we train ablation variants across all seven non-empty loss combinations for 40k steps. Figure~\ref{fig:res:loss-ablation} and Table~\ref{tab:res:loss-ablation} present these variants along with their normalized and absolute eval scores at $K{=}256$. The full objective achieves strong performance across all five metrics, demonstrating that the symmetric contrastive and generative losses do not conflict, but rather mutually reinforce one another.

\noindent
\begin{minipage}{\textwidth}
\begin{minipage}[t]{0.42\linewidth}
\vspace{0pt}
\centering
\includegraphics[width=\linewidth]{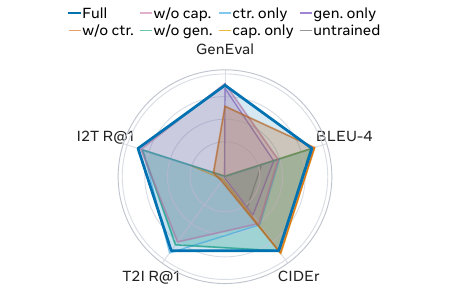}
\vspace{-9pt}
\captionof{figure}{\textbf{Eval for different training loss combinations (normalized scores).}}
\label{fig:res:loss-ablation}
\end{minipage}\hfill
\begin{minipage}[t]{0.56\linewidth}
\vspace{-6pt}
\centering
\captionof{table}{\textbf{Eval for different training loss combinations (absolute scores).}}
\label{tab:res:loss-ablation}
\vspace{2pt}
\small
\renewcommand{\arraystretch}{0.96}
\setlength{\tabcolsep}{3.4pt}
\begin{tabular}{lrrrrr}
\toprule
\multirow{2.5}{*}{\textbf{Losses}} & \multicolumn{2}{c}{\textbf{Retrieval}} & \multicolumn{2}{c}{\textbf{Captioning}} & \textbf{Generation} \\
\cmidrule(lr){2-3}\cmidrule(lr){4-5}\cmidrule(l){6-6}
& I2T & T2I & B@4 & CIDEr & GenEval \\
\midrule
\rowcolor{gray!12} Full & 52.2 & 56.3 & 40.2 & 137.5 & 0.327 \\
w/o contrastive & 13.6 & 6.7 & 40.7 & 139.2 & 0.297 \\
w/o captioning & 51.5 & 53.6 & 33.9 & 117.1 & 0.329 \\
w/o generation & 51.2 & 54.5 & 40.0 & 137.8 & 0.003 \\
ctr. only & 51.0 & 57.0 & 34.2 & 118.0 & 0.003 \\
cap. only & 39.4 & 4.0 & 40.8 & 139.7 & 0.004 \\
gen. only & 0.6 & 1.1 & 33.1 & 110.8 & 0.323 \\
untrained & 0.2 & 0.0 & 30.8 & 105.0 & 0.000 \\
\bottomrule
\end{tabular}
\end{minipage}
\end{minipage}

We further quantify how the three losses reinforce one another using a Shapley value decomposition~\citep{shapley1953value} over all eight coalitions. The attribution matrices and prefix sweeps in App.~\ref{app:loss-ablation} show that while each task is driven primarily by its matched loss, the other objectives generally provide complementary signals.

\FloatBarrier

\section{Conclusion}
\label{sec:conclusion}
We introduce FLAT, a representation pre-training framework that resamples images and text into 1D flexible-length, aligned, transmodal tokens. FLAT jointly optimizes a shared representation encoder alongside T2I and I2T decoders using bidirectional alignment and generation losses. This setup learns unified semantic representations not only useful for cross-modal retrieval but also as effective generative conditions for both decoders. Empirical results demonstrate that FLAT natively supports variable-length retrieval and synthesis, and with lightweight task-specific fine-tuning pushing T2I / I2T retrieval and generation performance to SOTA levels. Furthermore, qualitative analysis reveals that FLAT tokens enable  zero-shot linear interpolation, latent space arithmetic, and composed retrieval.

\bibliographystyle{assets/plainnat}
\bibliography{iclr2027_conference}

@inproceedings{radford2021clip,
  title     = {Learning Transferable Visual Models From Natural Language Supervision},
  author    = {Radford, Alec and Kim, Jong Wook and Hallacy, Chris and Ramesh, Aditya
               and Goh, Gabriel and Agarwal, Sandhini and Sastry, Girish and
               Askell, Amanda and Mishkin, Pamela and Clark, Jack and
               Krueger, Gretchen and Sutskever, Ilya},
  booktitle = {International Conference on Machine Learning (ICML)},
  year      = {2021}
}

@inproceedings{assran2023self,
  title={Self-supervised learning from images with a joint-embedding predictive architecture},
  author={Assran, Mahmoud and Duval, Quentin and Misra, Ishan and Bojanowski, Piotr and Vincent, Pascal and Rabbat, Michael and LeCun, Yann and Ballas, Nicolas},
  booktitle={2023 IEEE/CVF Conference on Computer Vision and Pattern Recognition (CVPR)},
  pages={15619--15629},
  year={2023},
  organization={IEEE}
}

@inproceedings{xiong2025gigatok,
  title={Gigatok: Scaling visual tokenizers to 3 billion parameters for autoregressive image generation},
  author={Xiong, Tianwei and Liew, Jun Hao and Huang, Zilong and Feng, Jiashi and Liu, Xihui},
  booktitle={2025 IEEE/CVF International Conference on Computer Vision (ICCV)},
  pages={18770--18780},
  year={2025},
  organization={IEEE}
}

@inproceedings{diao2025evev2,
  title={Evev2: Improved baselines for encoder-free vision-language models},
  author={Diao, Haiwen and Li, Xiaotong and Cui, Yufeng and Wang, Yueze and Deng, Haoge and Pan, Ting and Wang, Wenxuan and Lu, Huchuan and Wang, Xinlong},
  booktitle={2025 IEEE/CVF International Conference on Computer Vision (ICCV)},
  pages={21014--21025},
  year={2025},
  organization={IEEE}
}

@article{agrawal2024pixtral,
  title={Pixtral 12B},
  author={Agrawal, Pravesh and Antoniak, Szymon and Hanna, Emma Bou and Bout, Baptiste and Chaplot, Devendra and Chudnovsky, Jessica and Costa, Diogo and De Monicault, Baudouin and Garg, Saurabh and Gervet, Theophile and others},
  journal={arXiv preprint arXiv:2410.07073},
  year={2024}
}

@inproceedings{liu2023llava,
  title     = {Visual Instruction Tuning},
  author    = {Liu, Haotian and Li, Chunyuan and Wu, Qingyang and Lee, Yong Jae},
  booktitle = {Advances in Neural Information Processing Systems (NeurIPS)},
  year      = {2023}
}

@misc{team2026qwen3,
  title        = {{Qwen3.5-2B}},
  author       = {{Qwen Team}},
  year         = {2026},
  howpublished = {\url{https://huggingface.co/Qwen/Qwen3.5-2B}}
}

@inproceedings{yu2024titok,
  title     = {An Image is Worth 32 Tokens for Reconstruction and Generation},
  author    = {Yu, Qihang and Weber, Mark and Deng, Xueqing and Shen, Xiaohui and
               Cremers, Daniel and Chen, Liang-Chieh},
  booktitle = {Advances in Neural Information Processing Systems (NeurIPS)},
  year      = {2024}
}

@inproceedings{bachmann2025flextok,
  title     = {{FlexTok}: Resampling Images into 1D Token Sequences of Flexible Length},
  author    = {Bachmann, Roman and Allardice, Jesse and Mizrahi, David and
               Fini, Enrico and Kar, O{\u{g}}uzhan Fatih and Amirloo, Elmira and
               El-Nouby, Alaaeldin and Zamir, Amir and Dehghan, Afshin},
  booktitle = {International Conference on Machine Learning (ICML)},
  year      = {2025},
  note      = {arXiv:2502.13967}
}

@article{pan2025metaquery,
  title   = {Transfer between Modalities with {MetaQueries}},
  author  = {Pan, Xichen and Shukla, Satya Narayan and Singh, Aashu and
             Zhao, Zhuokai and Mishra, Shlok Kumar and Wang, Jialiang and
             Xu, Zhiyang and Chen, Jiuhai and Li, Kunpeng and Juefei-Xu, Felix
             and Hou, Ji and Xie, Saining},
  journal = {arXiv preprint arXiv:2504.06256},
  year    = {2025}
}

@inproceedings{rippel2014nested,
  title     = {Learning Ordered Representations with Nested Dropout},
  author    = {Rippel, Oren and Gelbart, Michael A. and Adams, Ryan P.},
  booktitle = {International Conference on Machine Learning (ICML)},
  year      = {2014}
}

@inproceedings{kusupati2022mrl,
  title     = {Matryoshka Representation Learning},
  author    = {Kusupati, Aditya and Bhatt, Gantavya and Rege, Aniket and
               Wallingford, Matthew and Sinha, Aditya and Ramanujan, Vivek and
               Howard-Snyder, William and Chen, Kaifeng and Kakade, Sham and
               Jain, Prateek and Farhadi, Ali},
  booktitle = {Advances in Neural Information Processing Systems (NeurIPS)},
  year      = {2022}
}

@inproceedings{xie2025sana,
  title     = {{SANA}: Efficient High-Resolution Image Synthesis with Linear
               Diffusion Transformers},
  author    = {Xie, Enze and Chen, Junsong and Chen, Junyu and Cai, Han and
               Tang, Haotian and Lin, Yujun and Zhang, Zhekai and Li, Muyang and
               Zhu, Ligeng and Lu, Yao and Han, Song},
  booktitle = {International Conference on Learning Representations (ICLR)},
  year      = {2025}
}

@inproceedings{esser2024sd3,
  title     = {Scaling Rectified Flow Transformers for High-Resolution Image
               Synthesis},
  author    = {Esser, Patrick and Kulal, Sumith and Blattmann, Andreas and
               Entezari, Rahim and M{\"u}ller, Jonas and Saini, Harry and
               Levi, Yam and Lorenz, Dominik and Sauer, Axel and Boesel, Frederic
               and Podell, Dustin and Dockhorn, Tim and English, Zion and
               Lacey, Kyle and Goodwin, Alex and Marek, Yannik and Rombach, Robin},
  booktitle = {International Conference on Machine Learning (ICML)},
  year      = {2024}
}

@inproceedings{leng2025repae,
  title     = {{REPA-E}: Unlocking {VAE} for End-to-End Tuning with Latent
               Diffusion Transformers},
  author    = {Leng, Xingjian and Singh, Jaskirat and Hou, Yunzhong and
               Xing, Zhenchang and Xie, Saining and Zheng, Liang},
  booktitle = {IEEE/CVF International Conference on Computer Vision (ICCV)},
  year      = {2025}
}

@inproceedings{qu2025tokenflow,
  title     = {{TokenFlow}: Unified Image Tokenizer for Multimodal Understanding
               and Generation},
  author    = {Qu, Liao and Zhang, Huichao and Liu, Yiheng and Wang, Xu and
               Jiang, Yi and Gao, Yiming and Ye, Hu and Du, Daniel K. and
               Yuan, Zehuan and Wu, Xinglong},
  booktitle = {IEEE/CVF Conference on Computer Vision and Pattern Recognition (CVPR)},
  year      = {2025}
}

@inproceedings{ma2025unitok,
  title   = {{UniTok}: A Unified Tokenizer for Visual Generation and Understanding},
  author  = {Ma, Chuofan and Jiang, Yi and Wu, Junfeng and Yang, Jihan and
             Yu, Xin and Yuan, Zehuan and Peng, Bingyue and Qi, Xiaojuan},
  booktitle = {Advances in Neural Information Processing Systems (NeurIPS)},
  year    = {2025},
  note      = {arXiv:2502.20321},
}

@article{zhang2025qlip,
  title   = {{QLIP}: Text-Aligned Visual Tokenization Unifies Auto-Regressive
             Multimodal Understanding and Generation},
  author  = {Zhao, Yue and Xue, Fuzhao and Reed, Scott and Fan, Linxi and
             Zhu, Yuke and Kautz, Jan and Yu, Zhiding and Krähenbühl, Philipp
             and Huang, De-An},
  journal = {arXiv preprint arXiv:2502.05178},
  year    = {2025}
}

@article{wu2024vilau,
  title   = {{VILA-U}: A Unified Foundation Model Integrating Visual
             Understanding and Generation},
  author  = {Wu, Yecheng and Zhang, Zhuoyang and Chen, Junyu and Tang, Haotian
             and Li, Dacheng and Fang, Yunhao and Zhu, Ligeng and Xie, Enze and
             Yin, Hongxu and Yi, Li and Han, Song and Lu, Yao},
  journal = {arXiv preprint arXiv:2409.04429},
  year    = {2024}
}

@inproceedings{team2024chameleon,
  title     = {Chameleon: Mixed-Modal Early-Fusion Foundation Models},
  author    = {{Chameleon Team}},
  booktitle = {arXiv preprint arXiv:2405.09818},
  year      = {2024}
}

@inproceedings{zhou2025transfusion,
  title     = {Transfusion: Predict the Next Token and Diffuse Images with One
               Multi-Modal Model},
  author    = {Zhou, Chunting and Yu, Lili and Babu, Arun and Tirumala, Kushal
               and Yasunaga, Michihiro and Shamis, Leonid and Kahn, Jacob and
               Ma, Xuezhe and Zettlemoyer, Luke and Levy, Omer},
  booktitle = {International Conference on Learning Representations (ICLR)},
  year      = {2025}
}

@inproceedings{li2023blip2,
  title     = {{BLIP-2}: Bootstrapping Language-Image Pre-training with Frozen
               Image Encoders and Large Language Models},
  author    = {Li, Junnan and Li, Dongxu and Savarese, Silvio and Hoi, Steven},
  booktitle = {International Conference on Machine Learning (ICML)},
  year      = {2023},
  note      = {arXiv:2301.12597},
}

@inproceedings{rombach2022high,
  title     = {High-Resolution Image Synthesis with Latent Diffusion Models},
  author    = {Rombach, Robin and Blattmann, Andreas and Lorenz, Dominik and
               Esser, Patrick and Ommer, Bj{"o}rn},
  booktitle = {IEEE/CVF Conference on Computer Vision and Pattern Recognition (CVPR)},
  pages     = {10684--10695},
  year      = {2022}
}

@inproceedings{podell2024sdxl,
  title     = {{SDXL}: Improving Latent Diffusion Models for High-Resolution
               Image Synthesis},
  author    = {Podell, Dustin and English, Zion and Lacey, Kyle and Blattmann,
               Andreas and Dockhorn, Tim and M{\"u}ller, Jonas and Penna, Joe
               and Rombach, Robin},
  booktitle = {International Conference on Learning Representations (ICLR)},
  year      = {2024},
  note      = {arXiv:2307.01952},
}

@inproceedings{liu2025crossflow,
  title     = {Flowing from Words to Pixels: A Noise-Free Framework for
               Cross-Modality Evolution},
  author    = {Liu, Qihao and Yin, Xi and Yuille, Alan and Brown, Andrew and
               Singh, Mannat},
  booktitle = {IEEE/CVF Conference on Computer Vision and Pattern Recognition
               (CVPR)},
  year      = {2025},
  note      = {arXiv:2412.15213},
}

@inproceedings{li2022blip,
  title     = {{BLIP}: Bootstrapping Language-Image Pre-training for Unified
               Vision-Language Understanding and Generation},
  author    = {Li, Junnan and Li, Dongxu and Xiong, Caiming and Hoi, Steven},
  booktitle = {International Conference on Machine Learning (ICML)},
  year      = {2022},
  note      = {arXiv:2201.12086},
}

@inproceedings{yu2022coca,
  title     = {{CoCa}: Contrastive Captioners are Image-Text Foundation Models},
  author    = {Yu, Jiahui and Wang, Zirui and Vasudevan, Vijay and Yeung, Legg and
               Seyedhosseini, Mojtaba and Wu, Yonghui},
  booktitle = {Transactions on Machine Learning Research},
  year      = {2022},
  note      = {arXiv:2205.01917}
}

@article{ramesh2022unclip,
  title   = {Hierarchical Text-Conditional Image Generation with {CLIP} Latents},
  author  = {Ramesh, Aditya and Dhariwal, Prafulla and Nichol, Alex and Chu, Casey and
             Chen, Mark},
  journal = {arXiv preprint arXiv:2204.06125},
  year    = {2022}
}

@inproceedings{jia2021scaling,
  title     = {Scaling Up Visual and Vision-Language Representation Learning With Noisy Text Supervision},
  author    = {Jia, Chao and Yang, Yinfei and Xia, Ye and Chen, Yi-Ting and
               Parekh, Zarana and Pham, Hieu and Le, Quoc V. and Sung, Yunhsuan
               and Li, Zhen and Duerig, Tom},
  booktitle = {International Conference on Machine Learning (ICML)},
  pages     = {4904--4916},
  year      = {2021}
}

@inproceedings{zhai2023sigmoid,
  title     = {Sigmoid Loss for Language Image Pre-Training},
  author    = {Zhai, Xiaohua and Mustafa, Basil and Kolesnikov, Alexander and Beyer, Lucas},
  booktitle = {IEEE/CVF International Conference on Computer Vision (ICCV)},
  pages     = {11975--11986},
  year      = {2023}
}

@article{sun2023eva,
  title   = {EVA-CLIP: Improved Training Techniques for CLIP at Scale},
  author  = {Sun, Quan and Fang, Yuxin and Wu, Ledell and Wang, Xinlong and Cao, Yue},
  journal = {arXiv preprint arXiv:2303.15389},
  year    = {2023}
}

@inproceedings{chen2024pixart,
  title     = {PixArt-$\alpha$: Fast Training of Diffusion Transformer for Photorealistic Text-to-Image Synthesis},
  author    = {Chen, Junsong and Yu, Jincheng and Ge, Chongjian and Yao, Lewei
               and Xie, Enze and Wu, Yue and Wang, Zhongdao and Kwok, James
               and Luo, Ping and Lu, Huchuan and Li, Zhenguo},
  booktitle = {International Conference on Learning Representations (ICLR)},
  year      = {2024}
}

@article{wu2024janus,
  title   = {Janus: Decoupling Visual Encoding for Unified Multimodal Understanding and Generation},
  author  = {Wu, Chengyue and Chen, Xiaokang and Wu, Zhiyu and Ma, Yiyang and
             Liu, Xingchao and Pan, Zizheng and Liu, Wen and Xie, Zhenda and
             Yu, Xingkai and Ruan, Chong and Luo, Ping},
  journal = {arXiv preprint arXiv:2410.13848},
  year    = {2024}
}

@inproceedings{li2023blip,
  title={Blip-2: Bootstrapping language-image pre-training with frozen image encoders and large language models},
  author={Li, Junnan and Li, Dongxu and Savarese, Silvio and Hoi, Steven},
  booktitle={International conference on machine learning},
  pages={19730--19742},
  year={2023},
  organization={PmLR}
}

@inproceedings{dong2024dreamllm,
  title={Dreamllm: Synergistic multimodal comprehension and creation},
  author={Dong, Runpei and Peng, Yuang and Qi, Zekun and Ge, Zheng and Yang, Jinrong and Zhao, Liang and Sun, Jianjian and Zhou, Hongyu and Wei, Haoran and Kong, Xiangwen and others},
  booktitle={International Conference on Learning Representations},
  volume={2024},
  pages={6666--6702},
  year={2024}
}

@article{zheng2023minigpt,
  title={Minigpt-5: Interleaved vision-and-language generation via generative vokens},
  author={Zheng, Kaizhi and He, Xuehai and Wang, Xin Eric},
  journal={arXiv preprint arXiv:2310.02239},
  year={2023}
}

@article{zhang2026openvision,
  title={OpenVision 3: A Family of Unified Visual Encoder for Both Understanding and Generation},
  author={Zhang, Letian and Ren, Sucheng and Liu, Yanqing and Li, Xianhang and Wang, Zeyu and Zhou, Yuyin and Yao, Huaxiu and Zheng, Zeyu and Nie, Weili and Liu, Guilin and others},
  journal={arXiv preprint arXiv:2601.15369},
  year={2026}
}

@article{fan2025prism,
  title={The prism hypothesis: Harmonizing semantic and pixel representations via unified autoencoding},
  author={Fan, Weichen and Diao, Haiwen and Wang, Quan and Lin, Dahua and Liu, Ziwei},
  journal={arXiv preprint arXiv:2512.19693},
  year={2025}
}

@inproceedings{yue2026uniflow,
  title={Uniflow: A unified pixel flow tokenizer for visual understanding and generation},
  author={Yue, Zhengrong and Zhang, Haiyu and Zeng, Xiangyu and Chen, Boyu and Wang, Chenting and Zhuang, Shaobin and Dong, Lu and Wang, Yi and Wang, Limin and Wang, Yali},
  booktitle={International Conference on Learning Representations},
  volume={2026},
  pages={84741--84772},
  year={2026}
}

@article{lin2025toklip,
  title={Toklip: Marry visual tokens to clip for multimodal comprehension and generation},
  author={Lin, Haokun and Wang, Teng and Ge, Yixiao and Ge, Yuying and Lu, Zhichao and Wei, Ying and Zhang, Qingfu and Sun, Zhenan and Shan, Ying},
  journal={arXiv preprint arXiv:2505.05422},
  year={2025}
}

@article{chen2025blip3,
  title={Blip3-o: A family of fully open unified multimodal models-architecture, training and dataset},
  author={Chen, Jiuhai and Xu, Zhiyang and Pan, Xichen and Hu, Yushi and Qin, Can and Goldstein, Tom and Huang, Lifu and Zhou, Tianyi and Xie, Saining and Savarese, Silvio and others},
  journal={arXiv preprint arXiv:2505.09568},
  year={2025}
}

@article{diao2026sensenova,
  title={Sensenova-u1: Unifying multimodal understanding and generation with neo-unify architecture},
  author={Diao, Haiwen and Wu, Penghao and Deng, Hanming and Wang, Jiahao and Bai, Shihao and Wu, Silei and Fan, Weichen and Ye, Wenjie and Tong, Wenwen and Fan, Xiangyu and others},
  journal={arXiv preprint arXiv:2605.12500},
  year={2026}
}

@article{huang2025ming,
  title={Ming-univision: Joint image understanding and generation with a unified continuous tokenizer},
  author={Huang, Ziyuan and Zheng, DanDan and Zou, Cheng and Liu, Rui and Wang, Xiaolong and Ji, Kaixiang and Chai, Weilong and Sun, Jianxin and Wang, Libin and Lv, Yongjie and others},
  journal={arXiv preprint arXiv:2510.06590},
  year={2025}
}

@article{xie2024show,
  title   = {Show-o: One Single Transformer to Unify Multimodal Understanding and Generation},
  author  = {Xie, Jinheng and Mao, Weijia and Bai, Zechen and Zhang, David Junhao
             and Wang, Weihao and Lin, Kevin Qinghong and Gu, Yuchao and
             Chen, Zhijie and Yang, Zhenheng and Shou, Mike Zheng},
  journal = {arXiv preprint arXiv:2408.12528},
  year    = {2024}
}

@inproceedings{chevalier2023autocompressor,
  title     = {Adapting Language Models to Compress Contexts},
  author    = {Chevalier, Alexis and Wettig, Alexander and Ajith, Anirudh and Chen, Danqi},
  booktitle = {Empirical Methods in Natural Language Processing (EMNLP)},
  pages     = {3829--3846},
  year      = {2023}
}

@inproceedings{karpathy2015deep,
  title={Deep visual-semantic alignments for generating image descriptions},
  author={Karpathy, Andrej and Fei-Fei, Li},
  booktitle={Proceedings of the IEEE conference on computer vision and pattern recognition},
  pages={3128--3137},
  year={2015}
}

@article{li2026dream,
  title   = {Unifying Contrastive and Generative Objectives for Visual Understanding and Text-to-Image Generation},
  author  = {Li, Chao and Li, Tianhong and Nuthalapati, Sai Vidyaranya and Chen, Hong-You and Shukla, Satya Narayan and Cheng, Jianpeng and Yang, Yonghuan and Xiao, Jun and Fan, Xiangjun and Singh, Aashu and Katabi, Dina and Mishra, Shlok Kumar},
  journal = {arXiv preprint arXiv:2603.02667},
  year    = {2026},
  url     = {https://arxiv.org/abs/2603.02667}
}

@inproceedings{wen2025csr,
  title     = {Beyond Matryoshka: Revisiting Sparse Coding for Adaptive Representation},
  author    = {Wen, Tiansheng and Wang, Yifei and Zeng, Zequn and Peng, Zhong and Su, Yudi and Liu, Xinyang and Chen, Bo and Liu, Hongwei and Jegelka, Stefanie and You, Chenyu},
  booktitle = {International Conference on Machine Learning (ICML)},
  year      = {2025},
  note      = {arXiv:2503.01776}
}

@inproceedings{ghosh2023geneval,
  title     = {GenEval: An Object-Focused Framework for Evaluating Text-to-Image Alignment},
  author    = {Ghosh, Dhruba and Hajishirzi, Hanna and Schmidt, Ludwig},
  booktitle = {Advances in Neural Information Processing Systems (NeurIPS)},
  year      = {2023},
  note      = {arXiv:2310.11513}
}

@incollection{shapley1953value,
  title     = {A Value for $n$-Person Games},
  author    = {Shapley, Lloyd S.},
  booktitle = {Contributions to the Theory of Games, Volume II},
  editor    = {Kuhn, Harold W. and Tucker, Albert W.},
  series    = {Annals of Mathematics Studies},
  volume    = {28},
  pages     = {307--318},
  publisher = {Princeton University Press},
  year      = {1953},
  doi       = {10.1515/9781400881970-018}
}

@inproceedings{liu2021cirr,
  title     = {Image Retrieval on Real-life Images with Pre-trained Vision-and-Language Models},
  author    = {Liu, Zheyuan and Rodriguez-Opazo, Cristian and Teney, Damien and Gould, Stephen},
  booktitle = {IEEE/CVF International Conference on Computer Vision (ICCV)},
  year      = {2021},
  note      = {arXiv:2108.04024}
}

@article{wang2024emu3,
  title   = {Emu3: Next-Token Prediction is All You Need},
  author  = {Wang, Xinlong and Zhang, Xiaosong and Luo, Zhengxiong and Sun, Quan and Cui, Yufeng and Wang, Jinsheng and Zhang, Fan and Wang, Yueze and Li, Zhen and Yu, Qiying and Zhao, Yingli and Ao, Yulong and Min, Xuebin and Li, Tao and Wu, Boya and Zhao, Bo and Zhang, Bowen and Wang, Liangdong and Liu, Guang and He, Zheqi and Yang, Xi and Liu, Jingjing and Lin, Yonghua and Huang, Tiejun and Wang, Zhongyuan},
  journal = {arXiv preprint arXiv:2409.18869},
  year    = {2024}
}

@article{ma2024janusflow,
  title   = {JanusFlow: Harmonizing Autoregression and Rectified Flow for Unified Multimodal Understanding and Generation},
  author  = {Ma, Yiyang and Liu, Xingchao and Chen, Xiaokang and Liu, Wen and Wu, Chengyue and Wu, Zhiyu and Pan, Zizheng and Xie, Zhenda and Zhang, Haowei and Yu, Xingkai and Zhao, Liang and Wang, Yisong and Liu, Jiaying and Ruan, Chong},
  journal = {arXiv preprint arXiv:2411.07975},
  year    = {2024}
}

@article{chen2025januspro,
  title   = {Janus-Pro: Unified Multimodal Understanding and Generation with Data and Model Scaling},
  author  = {Chen, Xiaokang and Wu, Zhiyu and Liu, Xingchao and Pan, Zizheng and Liu, Wen and Xie, Zhenda and Yu, Xingkai and Ruan, Chong},
  journal = {arXiv preprint arXiv:2501.17811},
  year    = {2025}
}

@inproceedings{jiang2024vlm2vecmmeb,
  title     = {VLM2Vec: Training Vision-Language Models for Massive Multimodal Embedding Tasks},
  author    = {Jiang, Ziyan and Meng, Rui and Yang, Xinyi and Yavuz, Semih and Zhou, Yingbo and Chen, Wenhu},
  booktitle = {International Conference on Learning Representations (ICLR)},
  year      = {2025},
  note      = {arXiv:2410.05160}
}

@article{tschannen2025siglip2,
  title   = {SigLIP 2: Multilingual Vision-Language Encoders with Improved Semantic Understanding, Localization, and Dense Features},
  author  = {Tschannen, Michael and Gritsenko, Alexey and Wang, Xiao and Naeem, Muhammad Ferjad and Alabdulmohsin, Ibrahim and Parthasarathy, Nikhil and Evans, Talfan and Beyer, Lucas and Xia, Ye and Mustafa, Basil and H{\'e}naff, Olivier and Harmsen, Jeremiah and Steiner, Andreas and Zhai, Xiaohua},
  journal = {arXiv preprint arXiv:2502.14786},
  year    = {2025}
}

@inproceedings{he2022mae,
  title     = {Masked Autoencoders Are Scalable Vision Learners},
  author    = {He, Kaiming and Chen, Xinlei and Xie, Saining and Li, Yanghao and Doll{\'a}r, Piotr and Girshick, Ross},
  booktitle = {IEEE/CVF Conference on Computer Vision and Pattern Recognition (CVPR)},
  year      = {2022},
  note      = {arXiv:2111.06377}
}

@article{liu2026tuna,
  title={Tuna-2: Pixel embeddings beat vision encoders for multimodal understanding and generation},
  author={Liu, Zhiheng and Ren, Weiming and Huang, Xiaoke and Chen, Shoufa and Li, Tianhong and Chen, Mengzhao and Ji, Yatai and He, Sen and Schult, Jonas and Zeng, Belinda and others},
  journal={arXiv preprint arXiv:2604.24763},
  year={2026}
}

@inproceedings{li2023mage,
  title={Mage: Masked generative encoder to unify representation learning and image synthesis},
  author={Li, Tianhong and Chang, Huiwen and Mishra, Shlok and Zhang, Han and Katabi, Dina and Krishnan, Dilip},
  booktitle={Proceedings of the IEEE/CVF conference on computer vision and pattern recognition},
  pages={2142--2152},
  year={2023}
}

@inproceedings{luo2023scdnet,
  title     = {Semantic-Conditional Diffusion Networks for Image Captioning},
  author    = {Luo, Jianjie and Li, Yehao and Pan, Yingwei and Yao, Ting and
               Feng, Jianlin and Chao, Hongyang and Mei, Tao},
  booktitle = {IEEE/CVF Conference on Computer Vision and Pattern Recognition (CVPR)},
  year      = {2023},
  note      = {arXiv:2212.03099}
}

@inproceedings{xiao2024florence2,
  title     = {{Florence-2}: Advancing a Unified Representation for a Variety of
               Vision Tasks},
  author    = {Xiao, Bin and Wu, Haiping and Xu, Weijian and Dai, Xiyang and
               Hu, Houdong and Lu, Yumao and Zeng, Michael and Liu, Ce and
               Yuan, Lu},
  booktitle = {IEEE/CVF Conference on Computer Vision and Pattern Recognition (CVPR)},
  year      = {2024},
  note      = {arXiv:2311.06242}
}

@inproceedings{liang2022mindthegap,
  title     = {Mind the Gap: Understanding the Modality Gap in Multi-modal Contrastive Representation Learning},
  author    = {Liang, Weixin and Zhang, Yuhui and Kwon, Yongchan and Yeung, Serena and Zou, James},
  booktitle = {Advances in Neural Information Processing Systems (NeurIPS)},
  year      = {2022},
  note      = {arXiv:2203.02053}
}

@inproceedings{caron2021emerging,
  title={Emerging properties in self-supervised vision transformers},
  author={Caron, Mathilde and Touvron, Hugo and Misra, Ishan and J{\'e}gou, Herv{\'e} and Mairal, Julien and Bojanowski, Piotr and Joulin, Armand},
  booktitle={2021 IEEE/CVF international conference on computer vision (ICCV)},
  pages={9630--9640},
  year={2021},
  organization={IEEE}
}

\clearpage
\beginappendix
\section{Overview}
\label{app:overview}

This appendix first details the implementation and evaluation protocols
(Appendix~\ref{app:config}), followed by zero-shot results from the pre-trained
checkpoint (Appendix~\ref{app:zeroshot}) and controlled task-only adaptation
comparisons (Appendix~\ref{app:pretraining-transfer}). We then provide expanded
qualitative galleries (Appendix~\ref{app:galleries}), analyze flexible-length
representations and sampling choices (Appendix~\ref{app:flexible-length}), examine
zero-shot transfer and latent composition (Appendix~\ref{app:crossdomain}), and
report loss and truncation ablations (Appendices~\ref{app:loss-ablation}
and~\ref{app:padding}).

\section{Implementation and evaluation details}
\label{app:config}

FLAT is jointly pre-trained on $65.7$M image--text pairs for $135$k steps on 8 nodes using the config in Table~\ref{tab:app:config}. Table~\ref{tab:app:finetune} summarizes the task-specific fine-tuning configs for
checkpoints evaluated in Sec.~\ref{sec:exp}

\begin{table}[!htbp]
\centering
\caption{FLAT model and pre-training configuration.}
\label{tab:app:config}
\small
\begin{tabular}{@{}p{0.30\textwidth}p{0.63\textwidth}@{}}
\toprule
\textbf{Component} & \textbf{Configuration} \\
\midrule
\multicolumn{2}{@{}l}{\textbf{Encoder and text decoder}} \\
Model      & Qwen3.5-2B, bfloat16, FlashAttention-2, gradient checkpointing \\
LoRA       & $r{=}16$, $\alpha{=}32$, dropout $0.05$; on Q/K/V/O and gate/up/down projections \\
Adapters   & Two adapters sharing this configuration: encoder and caption decoder \\
Base weights & Frozen; $11.4$M trainable across the adapter and register projections \\
\addlinespace
\multicolumn{2}{@{}l}{\textbf{Register sequence}} \\
Register tokens & $N=256$, latent dimension $d=64$ \\
Nested dropout  & $p=1.0$, $\mathcal{K}=\{1,4,16,64,256\}$ uniform; one shared $K$ per optimization step \\
Truncation      & Tail dropped, not null-padded, so a prefix is a shorter sequence (App.~\ref{app:padding}) \\
Minimum tokens  & 1 \\
Read-out head   & MLP with norm matching and latent normalization \\
Quantizer       & None; the representation stays continuous and is scored directly \\
\addlinespace
\multicolumn{2}{@{}l}{\textbf{Objectives}} \\
Loss weights          & Contrastive $1.0$; text generation $1.0$; image generation $1.0$ \\
Contrastive objective & InfoNCE, temperature $0.07$, cross-device negatives \\
Image conditioning    & Text registers at every step (\texttt{imagegen\_image\_cond\_prob}${=}0$) \\
Encoder instruction   & \emph{Represent the user's input.} \\
Caption instruction   & \emph{Describe the input.} \\
\addlinespace
\multicolumn{2}{@{}l}{\textbf{Image decoder}} \\
Model         & SANA-1600M at $512$ px, fully fine-tuned \\
Connector     & 16 Transformer layers, caption channels $2304$, input scale $2.3452$ \\
Flow matching & Flow objective, shift $3.0$, logit-normal $t$ with mean $0$ and std $1$ \\
CFG dropout   & $0.1$ \\
\addlinespace
\multicolumn{2}{@{}l}{\textbf{Data}} \\
Image resolution    & $512\times512$, giving a $16\times16$ grid of $256$ visual tokens \\
Maximum text length & 128 tokens \\
Training data       & $67$M image--text pairs \\
Approx. composition & Long-caption ($30$M), short-caption ($33$M), prompt-style ($4$M), and curated high-quality ($0.1$M) pairs \\
Data workers        & 20 per rank, prefetch factor 6 \\
\addlinespace
\multicolumn{2}{@{}l}{\textbf{Optimization}} \\
Optimizer      & AdamW, $\beta=(0.9, 0.999)$, weight decay $0.0$, gradient clip $1.0$ \\
Learning rate  & $1.0\times10^{-4}$; $500$-step linear warm-up then cosine decay to zero \\
Batch size     & $16$ per rank $\times$ $64$ ranks $\times$ 1 accumulation step $=1024$ global \\
Hardware       & 8 nodes $\times$ 8 H200 GPUs \\
Precision      & bfloat16 \\
Training steps & $135$k pretraining\\
\bottomrule
\end{tabular}
\end{table}

\begin{table*}[t]
\centering
\caption{\textbf{Task-specific fine-tuning configs.} Each column describes the config of the task-specific fine-tuned
checkpoint evaluated in Sec.~\ref{sec:exp}. All runs use AdamW, bfloat16, zero
weight decay, direct prefix truncation, and uniform sampling over
$\mathcal{K}=\{1,4,16,64,256\}$. Batch sizes are global; retrieval uses the
global batch as cross-device in-batch negatives.}
\label{tab:app:finetune}
\footnotesize
\renewcommand{\arraystretch}{1.12}
\setlength{\tabcolsep}{3pt}
\begin{tabular}{@{}>{\raggedright\arraybackslash}p{0.18\textwidth}
                    >{\raggedright\arraybackslash}p{0.185\textwidth}
                    >{\raggedright\arraybackslash}p{0.185\textwidth}
                    >{\raggedright\arraybackslash}p{0.17\textwidth}
                    >{\raggedright\arraybackslash}p{0.17\textwidth}@{}}
\toprule
\multirow{2.5}{=}{\textbf{Setting}}
& \multirow{2.5}{=}{\textbf{T2I synthesis}}
& \multirow{2.5}{=}{\textbf{I2T captioning}}
& \multicolumn{2}{c}{\textbf{Image--text retrieval}} \\
\cmidrule(lr){4-5}
& & & \textbf{COCO} & \textbf{Flickr30k} \\
\midrule
Training data
& 120K high-quality  image--text pairs
& COCO Karpathy train+restval (113,287 images)
& COCO Karpathy train+restval
& Flickr30k Karpathy train \\
\addlinespace[2pt]
Updated parameters
& Image decoder
& Text decoder LoRA
& Encoder LoRA, registers, and latent projection
& Encoder LoRA, registers, and latent projection \\
Objective
& T2I generation
& I2T generation
& T2I an I2T contrastive
& T2I an I2T contrastive \\
\addlinespace[2pt]
Checkpoint step
& 8,000
& 5,000
& 5,000
& 320 \\
Global batch
& 1,024
& 256
& 2,048
& 2,048 \\
Learning rate
& $4\!\times\!10^{-4}$
& $2\!\times\!10^{-4}$
& $1\!\times\!10^{-4}$
& $1\!\times\!10^{-4}$ \\
Warm-up steps
& 5,000
& 2,213
& 500
& 128 \\
\bottomrule
\end{tabular}
\end{table*}


\subsection{Evaluation protocol}
\label{app:eval}

\textbf{T2I Generation.} We evaluate text-to-image (T2I) generation using GenEval~\citep{ghosh2023geneval} across all $553$ prompts, generating $4$ images per prompt with $20$ sampling steps and a classifier-free guidance (CFG) scale of $4.5$. To ensure controlled comparisons across prefix lengths, the initial noise vector is sampled once per prompt and reused. We do not apply prompt rewriting at any stage.

\textbf{I2T Generation.} We evaluate image captioning (I2T) on the MS-COCO Karpathy test split using greedy decoding. Generated captions are evaluated via \texttt{pycocoevalcap}, reporting standard metrics: BLEU-4 (B@4), METEOR (M), ROUGE-L (R), CIDEr (C), and SPICE (S).
\begin{itemize}
    \item \textbf{BLEU-4 (B@4):} Measures $n$-gram precision (up to $4$-grams) between generated and reference captions, incorporating a brevity penalty to discourage overly short outputs.
    \item \textbf{METEOR (M):} Computes an $F$-score based on unigram precision and recall, accounting for word stemming, synonymy, and paraphrase alignment.
    \item \textbf{ROUGE-L (R):} Evaluates structural word-order similarity using the Longest Common Subsequence (LCS) between predicted captions and ground-truth references.
    \item \textbf{CIDEr (C):} Computes TF-IDF--weighted $n$-gram cosine similarity to reward visually descriptive content while penalizing generic terms.
    \item \textbf{SPICE (S):} Converts captions into semantic scene graphs (objects, attributes, and relations) and calculates the $F_1$-score over the extracted graph tuples.
\end{itemize}

\textbf{Retrieval.}We evaluate both text-to-image (T2I) and image-to-text (I2T) retrieval on the COCO Karpathy $5\mathrm{K}$ and Flickr30K Karpathy $1\mathrm{K}$ test splits, reporting direct retrieval recall without post-hoc reranking. 

\textbf{Composed Retrieval.} Following the MMEB protocol~\citep{jiang2024vlm2vecmmeb}, we evaluate on CIRR~\citep{liu2021cirr} using $1{,}000$ queries against a $1{,}000$-image gallery. We perform latent space arithmetic by converting image editing instructions into \textsc{add} and \textsc{remove} clauses. Specifically, let $\mathbf{z}_{\mathrm{ref}}$ denote the original image representation and $\mathbf{z}_0$ denote the representation of the neutral phrase \textit{"a photo"}. For an edit specified by added phrases $A$ and removed phrases $R$, the composite query representation is formulated as:$$\mathbf{z}_{q} = \mathbf{z}_{\mathrm{ref}} + \alpha \left[ \sum_{a \in A} (\mathbf{z}_a - \mathbf{z}_0) - \sum_{r \in R} (\mathbf{z}_r - \mathbf{z}_0) \right]$$After composition, we $\ell_2$-normalize each token representation and score the similarity with the late-interaction score defined in Eq.~\ref{eq:flat_sid_similarity}. We sweep $\alpha \in \{0.25, 0.5, 1, 1.5, 2, 3, 4, 6, 8\}$ for each $K$ on this diagnostic set. Table~\ref{tab:res:cirr} reports all metrics using the $\alpha$ selected by Recall@1.

\paragraph{Linear Probing.} For linear probing, we train a linear classification head on top of the frozen, concatenated prefix-$K$ token representations. We train on all $1{,}281{,}167$ ImageNet-1K training images for 20 epochs and evaluate performance on the $50{,}000$-image validation split.

\FloatBarrier

\section{Evaluation Results of the pre-trained checkpoint}
\label{app:zeroshot}

The jointly pre-trained checkpoint supports generation and retrieval without task-specific adaptation. We first evaluate generation across both modalities, followed by bidirectional retrieval.

\textbf{T2I Generation.} Table~\ref{tab:app:pretrain-geneval} reports GenEval results using the same prompt, sampling, and fixed-noise protocol as the fine-tuned checkpoint. Without task-specific tuning, the pre-trained model reaches an overall score of $0.71$, with performance scaling from $0.32$ using a single token to $0.71$ with $256$ tokens.

\begin{table}[!ht]
\centering
\caption{\textbf{GenEval for the pre-trained checkpoint.} No prompt rewriting is applied.}
\label{tab:app:pretrain-geneval}
\small
\setlength{\tabcolsep}{9pt}
\begin{tabular}{@{}lccccc@{}}
\toprule
$K$ & 1 & 4 & 16 & 64 & 256 \\
\midrule
GenEval & 0.317 & 0.613 & 0.703 & 0.701 & \textbf{0.711} \\
\bottomrule
\end{tabular}
\end{table}

\textbf{I2T Generation.}
We evaluate the pre-trained checkpoint directly on I2T captioning using the MS-COCO Karpathy test split. As shown in Table~\ref{tab:app:pretrain-caption}, performance steadily improves with larger prefix lengths, reaching 63.8 CIDEr and 16.3 SPICE at $K{=}256$. Although these absolute scores are lower than those of the fine-tuned checkpoint (Table~\ref{tab:res:caption}), this gap is largely driven by stylistic differences of training data: the pre-training corpus contains diverse descriptive text, whereas COCO favors short, literal captions. This distributional gap inherently penalizes overlap-based metrics, as semantically correct but longer or editorial descriptions introduce phrasing that diverges from compact reference captions.

\begin{figure}[!htbp]
\centering
\includegraphics[width=\textwidth]{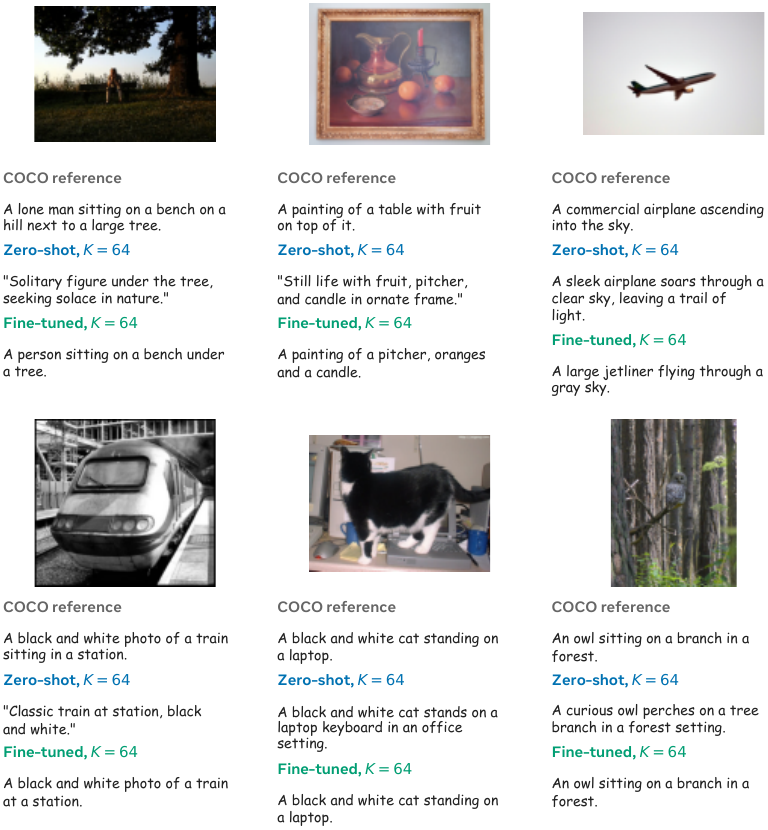}
\caption{\textbf{A comparison between the generated captions between pre-trained and fine-tuned checkpoints.}
Each example compares a COCO reference caption against outputs from the pre-trained and fine-tuned checkpoints. While zero-shot predictions from the pre-trained model reliably capture core entities and actions, fine-tuning aligns output length and phrasing with the COCO annotation style.}
\label{fig:app:zeroshot-caption}
\end{figure}

Figure~\ref{fig:app:zeroshot-caption} qualitatively illustrates this phenomenon. Zero-shot generations reliably capture the central entities and actions, but often express them as detailed descriptions or title-style captions. For instance, at $K{=}64$, 99.6\% of generated sequences properly terminate with an EOS token and average 10.8 words, yet 27.1\% begin with quotation marks characteristic of image titles. These results indicate that while pre-trained representations already encode rich visual semantics, fine-tuning primarily serves to align decoder length and stylistic tone with COCO expectations. Consequently, part of the metric gains observed after fine-tuning reflects output-format alignment rather than improved semantic understanding alone.

\begin{table}[tbp]
\centering
\caption{\textbf{Zero-shot I2T captioning evaluation for the pre-trained checkpoint.}
Greedy decoding is used to generate the captions.}
\label{tab:app:pretrain-caption}
\small
\setlength{\tabcolsep}{7pt}
\begin{tabular}{@{}lrrrrr@{}}
\toprule
$K$ & B@4 & M & R & C & S \\
\midrule
1   & 11.0 & 17.9 & 37.2 & 46.1 & 12.5 \\
4   & 13.4 & 20.2 & 39.8 & 57.2 & 14.6 \\
16  & 14.5 & 21.3 & 41.3 & 62.1 & 15.7 \\
64  & 14.9 & 21.6 & 41.6 & 63.5 & 16.2 \\
256 & \textbf{15.0} & \textbf{21.7} & \textbf{41.7} & \textbf{63.8} & \textbf{16.3} \\
\bottomrule
\end{tabular}
\end{table}

\FloatBarrier

\textbf{Retrieval.}
Table~\ref{tab:app:pretrain-retrieval} evaluates the pre-trained checkpoint on the full MS-COCO and Flickr30K Karpathy test splits without retrieval-specific fine-tuning. Even in this zero-shot setting, the learned representations effectively support bidirectional retrieval. However, zero-shot performance degrades slightly as the prefix length $K$ increases. Task-specific fine-tuning substantially mitigates this sensitivity to $K$ (Table~\ref{tab:res:ret}).

\begin{table}[tbp]
\centering
\caption{\textbf{Zero-shot retrieval for the pre-trained checkpoint.} Results are recall (\%) on the COCO Karpathy 5k and Flickr30k
Karpathy 1k test splits.}
\label{tab:app:pretrain-retrieval}
\small
\setlength{\tabcolsep}{8.5pt}
\begin{tabular}{@{}lrrrrrr@{}}
\toprule
& \multicolumn{3}{c}{\textbf{Image $\rightarrow$ text}}
& \multicolumn{3}{c}{\textbf{Text $\rightarrow$ image}} \\
\cmidrule(lr){2-4}\cmidrule(lr){5-7}
$K$ & R@1 & R@5 & R@10 & R@1 & R@5 & R@10 \\
\midrule
\multicolumn{7}{@{}l}{\emph{MS-COCO Karpathy 5k}}\\
1   & \textbf{44.90} & \textbf{69.10} & \textbf{78.54} & \textbf{39.70} & \textbf{64.59} & \textbf{74.51} \\
4   & 44.48 & 68.26 & 78.08 & 37.85 & 62.31 & 72.08 \\
16  & 41.94 & 66.96 & 76.32 & 36.32 & 61.02 & 70.84 \\
64  & 39.96 & 65.02 & 74.64 & 34.33 & 59.07 & 69.37 \\
256 & 35.48 & 60.04 & 70.64 & 31.45 & 55.87 & 66.57 \\
\addlinespace[1.5pt]
\multicolumn{7}{@{}l}{\emph{Flickr30k Karpathy 1k}}\\
1   & \textbf{64.10} & \textbf{89.70} & \textbf{94.90} & \textbf{64.40} & \textbf{86.04} & \textbf{91.16} \\
4   & 63.50 & 89.10 & 94.60 & 61.52 & 84.62 & 89.76 \\
16  & 61.00 & 89.00 & 94.60 & 60.44 & 83.98 & 89.30 \\
64  & 59.30 & 88.10 & 93.50 & 58.68 & 83.00 & 88.64 \\
256 & 55.70 & 84.30 & 91.90 & 55.74 & 81.30 & 87.20 \\
\bottomrule
\end{tabular}
\end{table}
\FloatBarrier

\section{Effect of FLAT pre-training on task adaptation}
\label{app:pretraining-transfer}

We compare task-specific adaptation with and without FLAT pre-training. The
task-only controls use the same architecture, training data, optimization
budget, and evaluation protocol as the corresponding fine-tuned FLAT models.
They retain the public backbone initialization but begin with randomly
initialized FLAT-specific parameters and receive no joint FLAT pre-training.
Table~\ref{tab:app:task-only} reports the complete prefix-length sweep. For T2I
synthesis, the task-only model remains near $0.60$ GenEval across all prefix
lengths. FLAT pre-training improves GenEval by $0.16$--$0.24$ for $K\geq4$ and
reaches $0.83$ at $K{=}16$ and $K{=}64$. Although the task-only model is higher
at $K{=}1$, its performance does not increase with additional tokens. In
contrast, the pre-trained initialization gains $0.34$ from $K{=}1$ to
$K{=}16$, showing that pre-training enables the image decoder to exploit wider
prefixes for compositional generation. Pre-training also improves captioning by
$9.0$--$10.9$ BLEU-4 and $31.9$--$41.2$ CIDEr across $K$. Its effect on
retrieval is smaller and direction-dependent: image-to-text R@1 improves by
$0.6$--$1.3$ points, while the task-only model is $0.4$--$0.9$ points higher
for text-to-image R@1. These results show that the generative tasks benefit most
from the representation learned during joint pre-training, whereas in-domain
contrastive adaptation learns most of the retrieval interface directly.

\begin{table}[htbp]
\centering
\caption{\textbf{Task-specific adaptation with and without FLAT pre-training.}
T2I uses matched $8$k-step recipes on 120K image--text pairs; captioning and
retrieval use matched $5$k-step recipes on the COCO Karpathy train+restval
split. T2I is evaluated on GenEval without prompt rewriting. Captioning and
retrieval are evaluated on the Karpathy 5k test split, with full galleries for
retrieval.}
\label{tab:app:task-only}
\small
\setlength{\tabcolsep}{6pt}
\begin{tabular}{llrrrrr}
\toprule
Metric & Initialization & $K{=}1$ & $K{=}4$ & $K{=}16$ & $K{=}64$ & $K{=}256$ \\
\midrule
\multirow{2}{*}{T2I GenEval}
& Task-only        & \textbf{0.60} & 0.61 & 0.61 & 0.60 & 0.59 \\
& FLAT pre-trained & 0.49 & \textbf{0.77} & \textbf{0.83} & \textbf{0.83} & \textbf{0.82} \\
\addlinespace[2pt]
\multirow{2}{*}{Captioning B@4}
& Task-only        & 24.4 & 27.3 & 30.6 & 31.0 & 30.8 \\
& FLAT pre-trained & \textbf{35.3} & \textbf{38.1} & \textbf{39.6} & \textbf{40.2} & \textbf{40.5} \\
\addlinespace[2pt]
\multirow{2}{*}{Captioning CIDEr}
& Task-only        & 80.8 & 91.5 & 104.3 & 105.3 & 105.0 \\
& FLAT pre-trained & \textbf{122.0} & \textbf{131.0} & \textbf{136.2} & \textbf{138.1} & \textbf{138.6} \\
\addlinespace[2pt]
\multirow{2}{*}{I2T retrieval R@1}
& Task-only        & 62.44 & 62.38 & 62.28 & 62.72 & 62.24 \\
& FLAT pre-trained & \textbf{63.00} & \textbf{63.40} & \textbf{63.30} & \textbf{63.32} & \textbf{63.54} \\
\addlinespace[2pt]
\multirow{2}{*}{T2I retrieval R@1}
& Task-only        & \textbf{48.62} & \textbf{48.54} & \textbf{48.49} & \textbf{48.53} & \textbf{48.43} \\
& FLAT pre-trained & 47.76 & 47.93 & 47.94 & 47.73 & 47.98 \\
\bottomrule
\end{tabular}
\end{table}
\FloatBarrier

\section{Qualitative results for FLAT}
\label{app:galleries}

\begin{figure}[htbp]
\centering
\includegraphics[width=\textwidth]{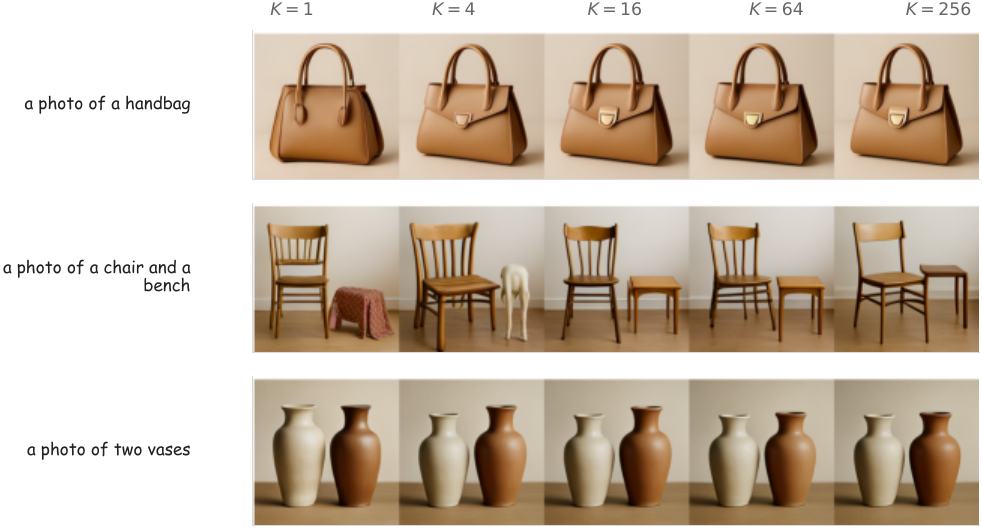}
\caption{\textbf{Text-to-image gallery, part I.} Samples are drawn from the fine-tuned
checkpoint across the full prefix-length ladder. The initial noise is fixed
within each row. These examples cover a single object, two-object composition,
and counting.}
\label{fig:app:t2i-gallery-1}
\end{figure}

\begin{figure}[htbp]
\centering
\includegraphics[width=\textwidth]{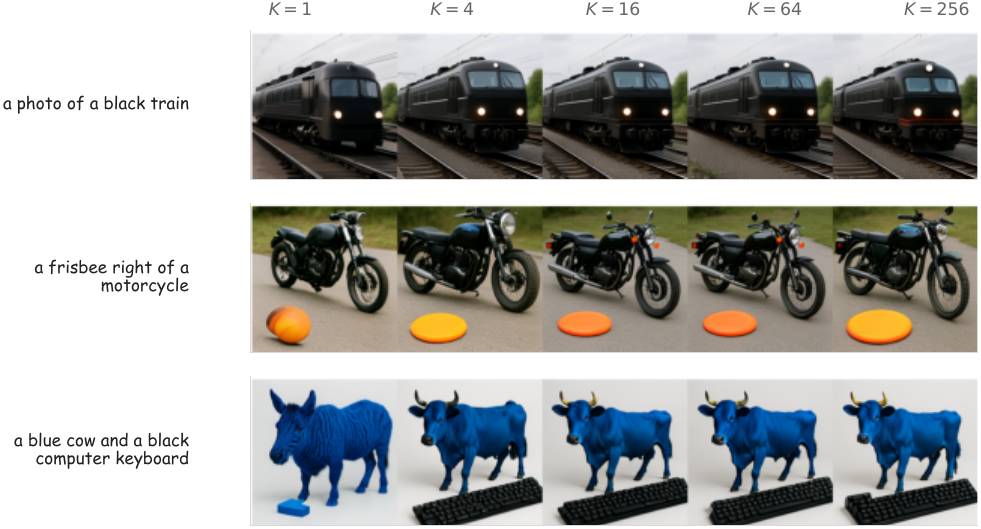}
\caption{\textbf{Text-to-image gallery, part II.} Additional samples covering
color, spatial relations, and attribute--object binding, with fixed initial
noise across prefix lengths.}
\label{fig:app:t2i-gallery-2}
\end{figure}

\begin{figure}[htbp]
\centering
\includegraphics[width=\textwidth]{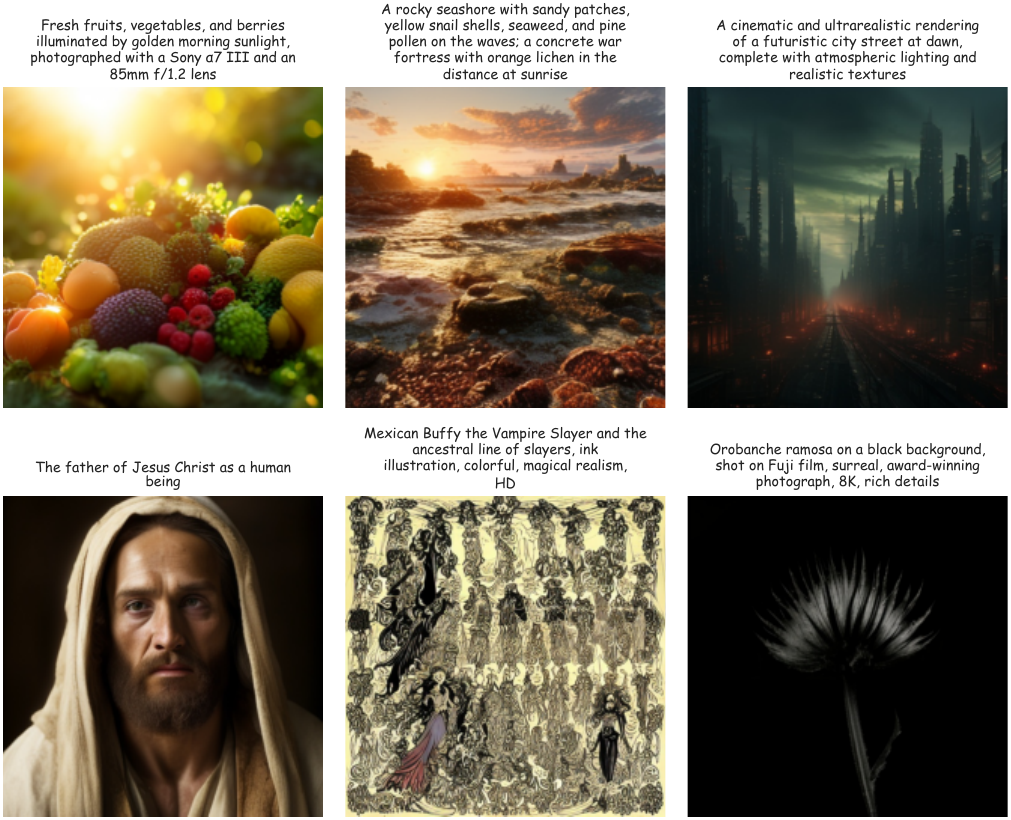}
\caption{\textbf{Text-to-image gallery with free-form prompts.} Samples at
$K{=}64$ for longer prompts from MJHQ-30K, spanning photographic and
illustrative styles across diverse subjects and scenes.}
\label{fig:app:t2i-gallery-complex}
\end{figure}

\begin{figure}[t]
\centering
\includegraphics[width=\textwidth]{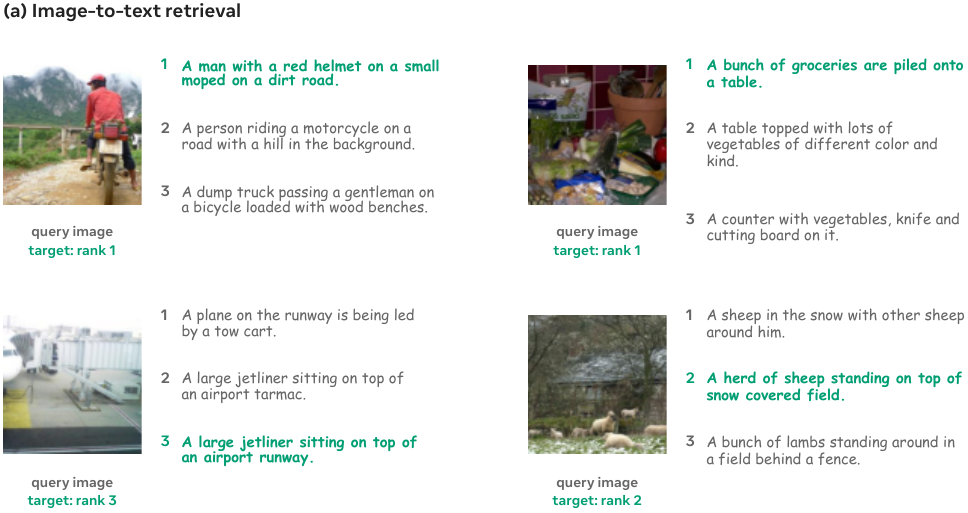}
\vspace{4pt}

\includegraphics[width=\textwidth]{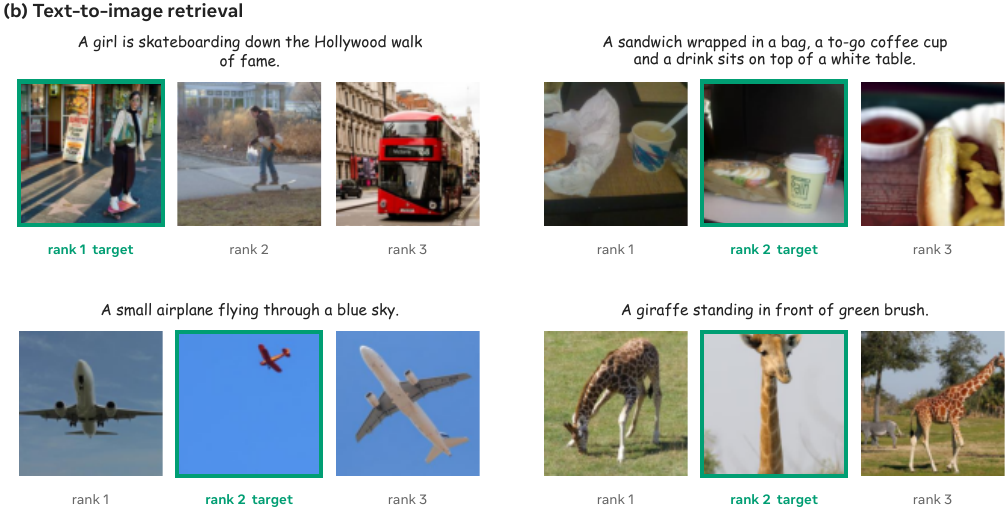}
\caption{\textbf{Bidirectional retrieval galleries.} Image-to-text (top) and text-to-image (bottom) retrieval results at $K{=}16$ on MS-COCO. A green box highlights the ground-truth target, while remaining images/texts show top-ranked retrieved candidates. Cases where the ground truth appears at rank 2 or 3 illustrate near-miss failures, where leading alternatives remain semantically consistent with the query.}
\label{fig:app:retrieval-gallery}
\end{figure}

\begin{figure}[t]
\centering
\includegraphics[width=\textwidth]{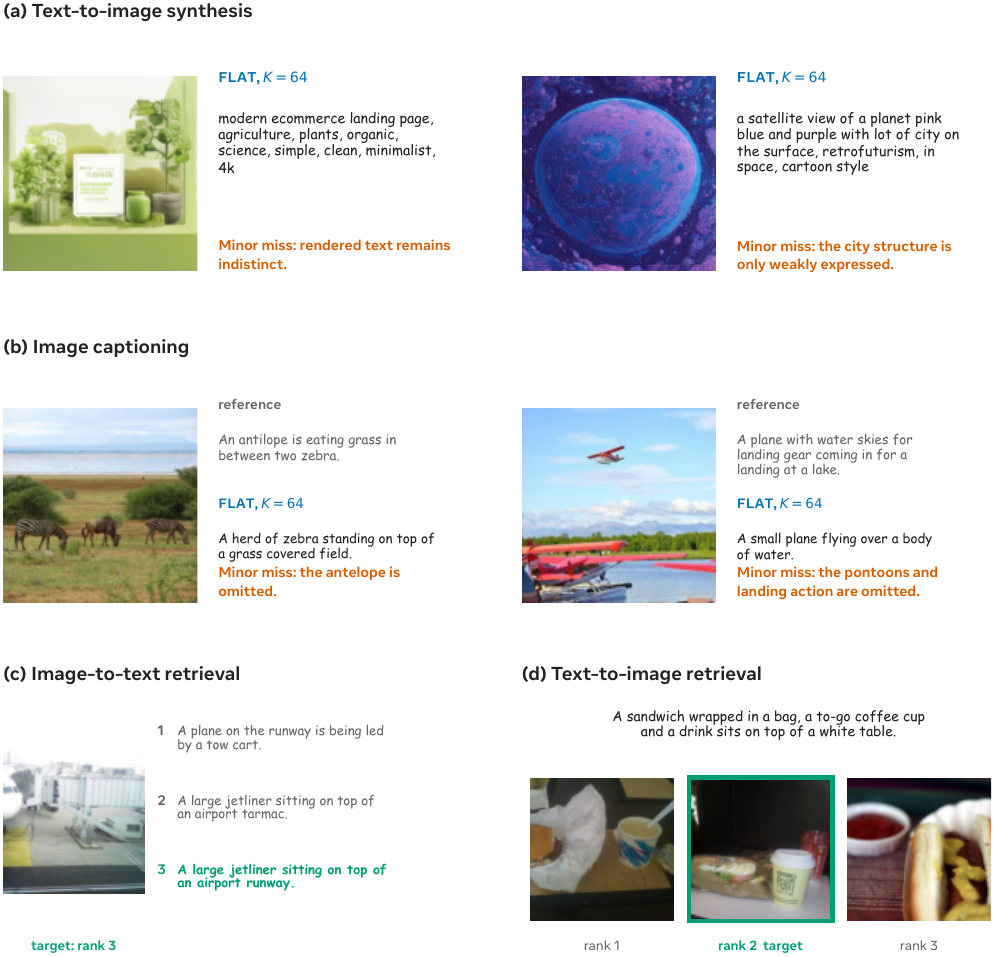}
\caption{\textbf{Representative failure cases across tasks.}
For text-to-image synthesis (top), the overall subject matter and style are preserved, though fine-grained layout or typography exhibits minor imperfections. For captioning (middle), the primary scene is described accurately, but secondary objects or actions are occasionally omitted. In bidirectional retrieval, ground-truth targets consistently rank near the top, preceded only by semantically close alternatives.}
\label{fig:app:failures}
\end{figure}

The following galleries present a qualitative evaluation of FLAT. Text-to-image samples share the same initial noise across prefix lengths, while retrieval evaluations utilize task-specific checkpoints and full galleries as described in Table~\ref{tab:app:finetune}. Figures~\ref{fig:app:t2i-gallery-1} and~\ref{fig:app:t2i-gallery-2} evaluate performance on controlled compositional prompts, whereas Figure~\ref{fig:app:t2i-gallery-complex} extends this comparison to longer, free-form prompts. Figure~\ref{fig:app:retrieval-gallery} illustrates bidirectional retrieval results, and Figure~\ref{fig:app:failures} highlights representative failure cases across the primary tasks.

\FloatBarrier

\section{Flexible-length representation analysis}
\label{app:flexible-length}

During training, FLAT samples $K$ only from a list
$\mathcal{K}=\{1,4,16,64,256\}$. We examine
whether the other choice of K can be flexibly applied at test time.

\subsection{Generalization to unseen prefix lengths}
\label{app:any-k}

\begin{figure}[!htbp]
\centering
\includegraphics[width=\textwidth]{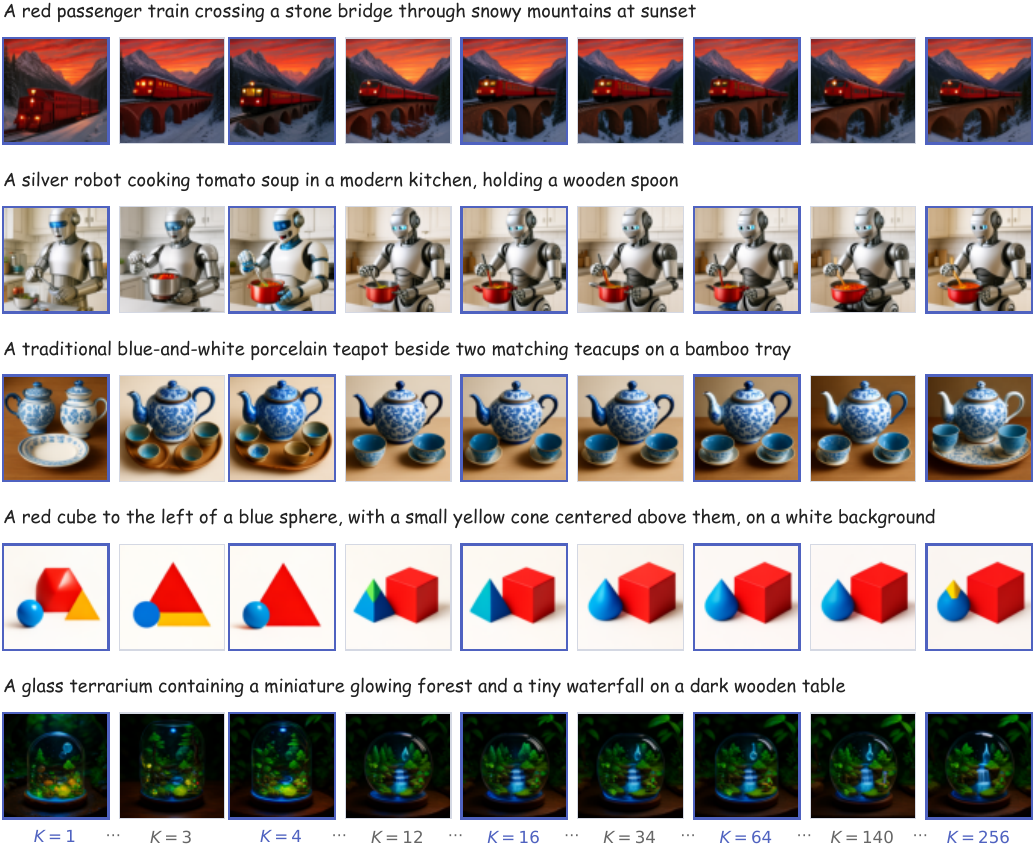}
\caption{\textbf{Generation at trained and unseen prefix lengths.} Blue borders indicate sequence lengths sampled during training, whereas gray borders denote unseen intermediate lengths. All columns share the same prompt, sampling configuration, and initial noise vector.}
\label{fig:app:any-k}
\end{figure}

In short, direct prefix truncation enables inference at arbitrary integer sequence lengths within 256. Figure~\ref{fig:app:any-k} illustrates outputs generated from the same prompt and initial noise across both trained lengths and unseen intermediate lengths. Generations at unseen prefix lengths remain coherent and evolve smoothly with $K$.

\subsection{Prefix-dependent compute}
Because the representation encoder is causal, the hidden states of the first $K$ registers are independent of subsequent register positions. At inference time, we can therefore process only the first $K$ registers rather than encoding all $N=256$ registers and truncating post-hoc. Table~\ref{tab:app:encoder-flops} reports single-batch inference FLOPs for the encoder and both decoders, counting each multiply--add operation as two FLOPs.  Reducing $K$ from $256$ to $1$ reduces encoder and I2T decoder compute by $50.0\%$ and $71.1\%$, respectively. In contrast, T2I compute decreases by $13.8\%$, as its cost is overwhelmingly dominated by spatial denoising and VAE decoding rather than conditioning length.

\begin{center}
\begin{minipage}{\textwidth}
\centering
\captionof{table}{\textbf{Encoder and decoder compute under prefix truncation.}
Absolute cost is reported in TFLOPs; relative cost normalizes each component to
its $K{=}256$ setting.}
\label{tab:app:encoder-flops}
\scriptsize
\setlength{\tabcolsep}{5pt}
\begin{tabular}{@{}r rr rr rr@{}}
\toprule
$K$ & \multicolumn{2}{c}{Encoder} & \multicolumn{2}{c}{I2T decoder} & \multicolumn{2}{c}{T2I decoder} \\
\cmidrule(lr){2-3}\cmidrule(lr){4-5}\cmidrule(l){6-7}
& TFLOPs & Relative & TFLOPs & Relative & TFLOPs & Relative \\
\midrule
1   & 0.714 & 50.0\% & 0.289 & 28.9\% & 29.964 & 86.2\% \\
4   & 0.723 & 50.6\% & 0.298 & 29.7\% & 30.020 & 86.4\% \\
16  & 0.756 & 52.9\% & 0.331 & 33.1\% & 30.245 & 87.1\% \\
64  & 0.891 & 62.3\% & 0.465 & 46.4\% & 31.144 & 89.6\% \\
256 & 1.430 & 100.0\% & 1.002 & 100.0\% & 34.741 & 100.0\% \\
\bottomrule
\end{tabular}
\end{minipage}
\end{center}

\FloatBarrier

\subsection{Prefix-length sampling strategies}
\label{sec:app:k-sampling}

We sample the prefix $K$ from the ladder $\mathcal{K}=\{1,4,16,64,256\}$ using
three strategies: 1)~\textbf{Uniform} assigns equal probability to every $K$;
2)~\textbf{Geometric} samples $K_r$ with $p(K_r)\propto\beta^r$, where
$\beta=1.3$; and 3)~\textbf{Replay} spends half of training on shorter prefixes
and the other half on the full prefix.

\begin{figure}[!htbp]
\centering
\includegraphics[width=\textwidth]{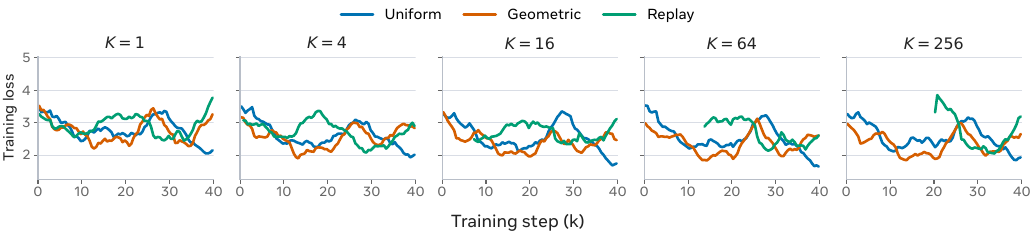}
\vspace{-3pt}
\caption{\textbf{Training loss under different prefix-length sampling strategies.}
Training lasts $40$k steps.}
\label{fig:app:k-sampling-loss}
\end{figure}

Figure~\ref{fig:app:k-sampling-loss} compares the training loss through $40$k
steps.
Uniform and Geometric sample every $K$ throughout training, whereas Replay
progressively unlocks longer prefixes.

\vspace{-7pt}
\begin{table}[!htbp]
\centering

\scriptsize
\renewcommand{\arraystretch}{0.92}
\setlength{\tabcolsep}{5.2pt}
\begin{tabular}{@{}llccccc@{}}
\toprule
Metric & Sampling & $K{=}1$ & $K{=}4$ & $K{=}16$ & $K{=}64$ & $K{=}256$ \\
\midrule
\multirow{3}{*}{GenEval ($\times100$) $\uparrow$}
& Uniform   & \textbf{33.0} & 32.5 & 32.4 & 31.9 & 32.7 \\
& Geometric & 26.0 & \textbf{43.2} & \textbf{49.9} & \textbf{49.9} & \textbf{50.2} \\
& Replay    & 32.2 & 32.6 & 32.2 & 32.0 & 32.3 \\
\cmidrule(lr){1-7}
\multirow{3}{*}{I2T R@1 $\uparrow$}
& Uniform   & 57.8 & 53.9 & 54.3 & 52.9 & \textbf{52.2} \\
& Geometric & 56.1 & 52.6 & 53.1 & 52.2 & 51.8 \\
& Replay    & \textbf{59.1} & \textbf{55.7} & \textbf{55.8} & \textbf{54.5} & 52.1 \\
\cmidrule(lr){1-7}
\multirow{3}{*}{T2I R@1 $\uparrow$}
& Uniform   & \textbf{62.4} & \textbf{60.1} & 58.4 & 56.7 & 56.3 \\
& Geometric & 60.9 & 56.9 & 56.6 & 55.8 & 55.5 \\
& Replay    & 62.2 & \textbf{60.1} & \textbf{59.7} & \textbf{59.2} & \textbf{59.1} \\
\cmidrule(lr){1-7}
\multirow{3}{*}{CIDEr $\uparrow$}
& Uniform   & 122.7 & 130.4 & 136.8 & \textbf{137.6} & 137.5 \\
& Geometric & \textbf{123.0} & 129.6 & \textbf{137.0} & 137.3 & \textbf{138.3} \\
& Replay    & 121.5 & \textbf{131.3} & 136.5 & 137.5 & 137.5 \\
\bottomrule
\end{tabular}
\caption{\textbf{Downstream performance under different prefix-length sampling
strategies.} }
\label{tab:app:k-sampling-ft}
\end{table}

Table~\ref{tab:app:k-sampling-ft} evaluates matched $40$k-step checkpoints. Geometric favors generation for $K\geq4$, Replay favors retrieval. However, no single strategy is universally optimal across all metrics and values of K.
\FloatBarrier

\subsection{Latent interpolation}
\label{app:interp}

\begin{figure}[!htbp]
\centering
\includegraphics[width=0.92\textwidth]{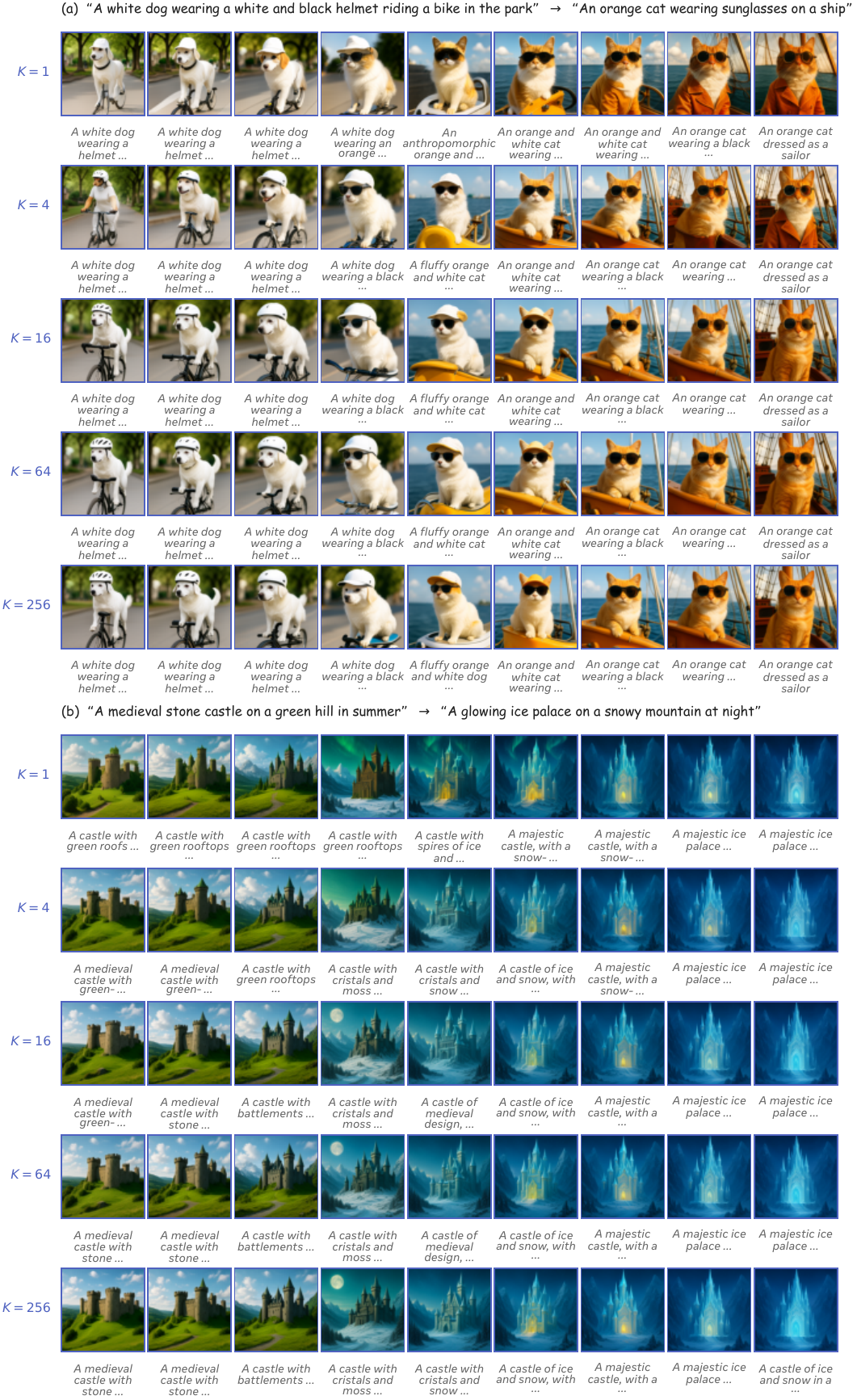}
\caption{\textbf{Latent walks across prefix lengths.} Two image--text 
pairs are interpolated at $K\in\{1,4,16,64,256\}$. Italic text under every
image is produced by the I2T decoder.}
\label{fig:app:interp}
\end{figure}

\begin{figure}[!htbp]
\centering
\vspace{-2mm}
\includegraphics[width=0.95\textwidth]{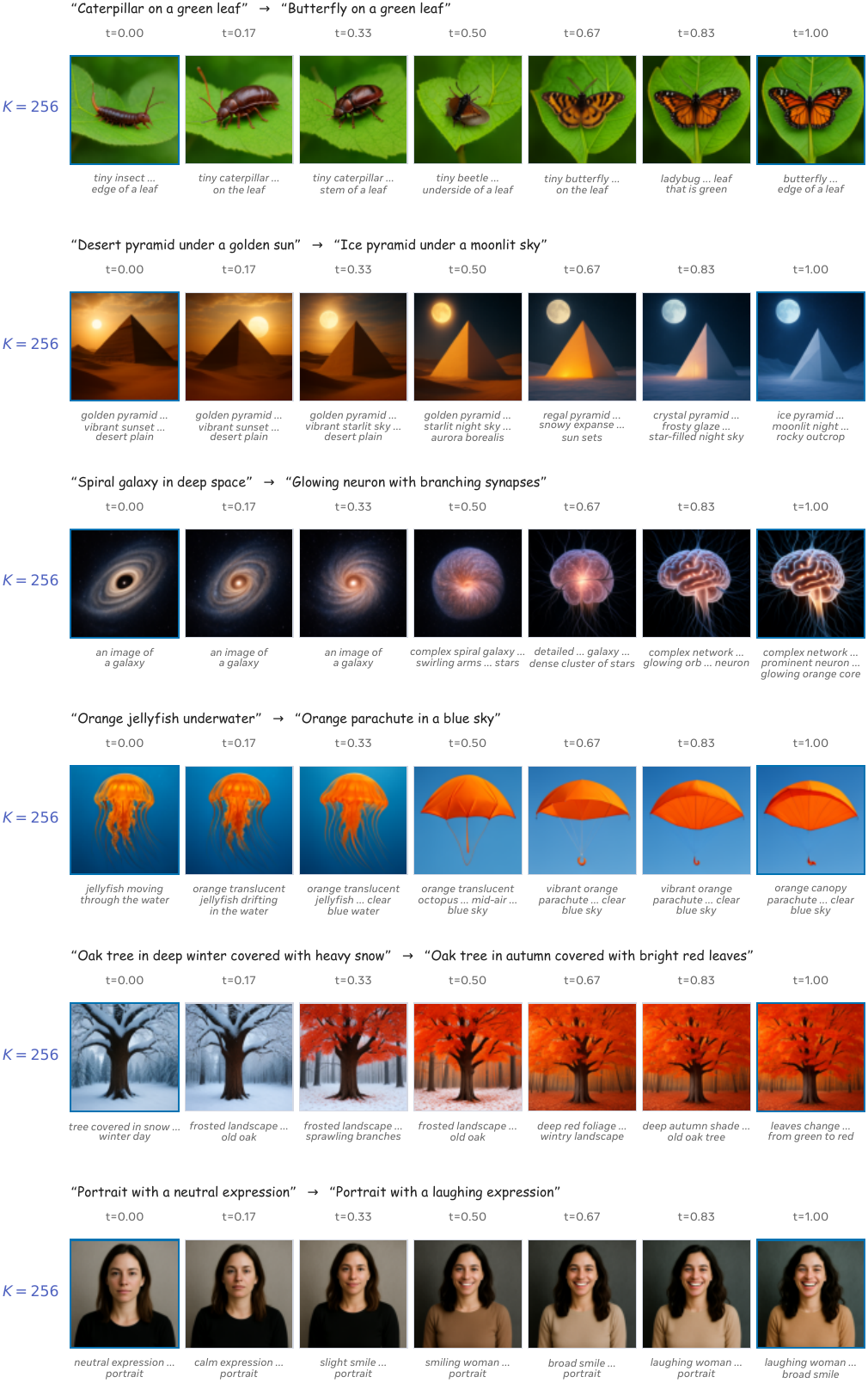}
\caption{\textbf{Additional latent walks at $K{=}256$.} Blue borders mark the
two endpoints. Italic text under every image is produced by the I2T decoder.}
\label{fig:app:interp-more}
\end{figure}

Figures~\ref{fig:app:interp} and~\ref{fig:app:interp-more} examine latent
interpolation through both decoders. Across five prefix lengths,
Figure~\ref{fig:app:interp} shows that longer prefixes transition multiple
attributes smoothly, while shorter prefixes yield coarser blends.
Figure~\ref{fig:app:interp-more} adds six $K{=}256$ walks spanning identity,
appearance, structure, context, and season; captions are condensed to keywords.

\FloatBarrier

\section{Zero-shot transfer and latent composition}
\label{app:crossdomain}

In this section, we demonstrate that the pre-trained FLAT representation natively supports zero-shot multilingual text understanding, multi-frame video understanding, and composed retrieval.

\subsection{Multilingual text and emoji understanding}
\begin{figure}[!htbp]
\centering
\includegraphics[width=0.86\textwidth]{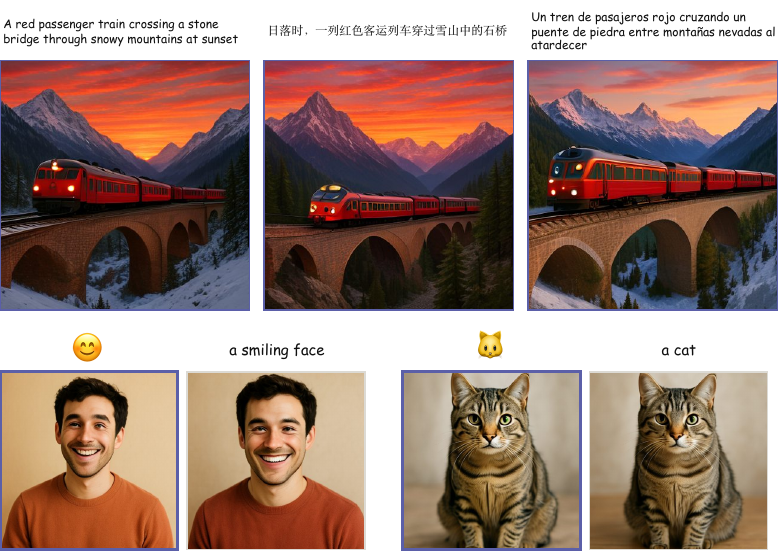}
\caption{\textbf{Zero-shot multilingual text and emoji conditioned image generation.} Although trained exclusively on English data, FLAT generalizes its understanding to non-English languages and emojis.}
\label{fig:res:multi}
\end{figure}

We first show that the representation encoder inherits multilingual capabilities from its LLM backbone. Consequently, non-English prompts occupy a similar representation space to English, despite the model being trained exclusively on English data. Figure~\ref{fig:res:multi} demonstrates that French and Chinese prompts decode into images consistent with their English counterparts, while emoji sequences reliably decode into their corresponding visual referents.

\FloatBarrier

\subsection{Multi-frame video understanding}
\begin{figure}[!htbp]
\centering
\includegraphics[width=\textwidth]{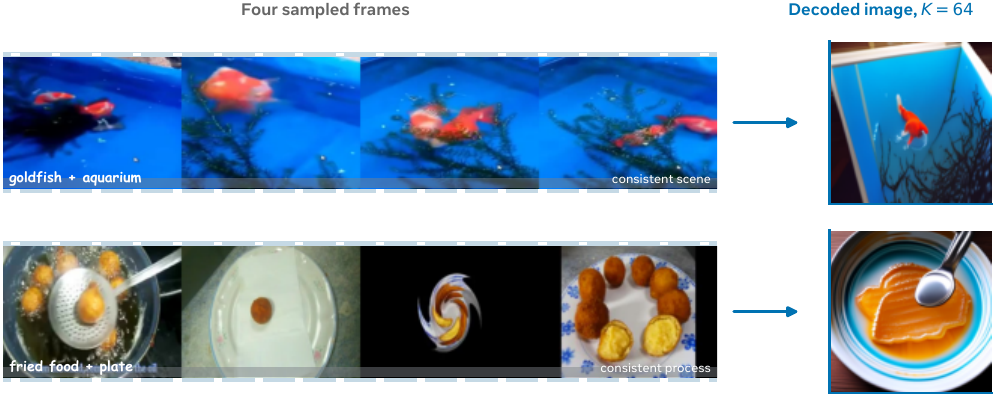}
\caption{\textbf{Zero-shot multi-frame video semantic aggregation.} Each row jointly
encodes 4 sampled frames and decodes one image at $K{=}64$. The generated outputs preserve persistent objects and scene context across the sampled frames.}
\label{fig:res:video}
\end{figure}

We next evaluate FLAT's ability to process multi-frame video inputs despite being trained exclusively on static images. Specifically, we jointly encode $4$ sampled frames using the representation encoder and decode the resulting sequence via the image generation head. Figure~\ref{fig:res:video} shows examples from an aquarium scene and a cooking sequence. At $K{=}64$, the generated outputs preserve persistent objects and scene context while integrating information across the sampled frames. These results demonstrate that the FLAT representation naturally aggregates multi-frame context into a unified, thumbnail-like visual output.

\FloatBarrier

\subsection{Composed retrieval}
\label{app:cirr}

\begin{figure}[!htbp]
\centering
\includegraphics[width=0.74\textwidth]{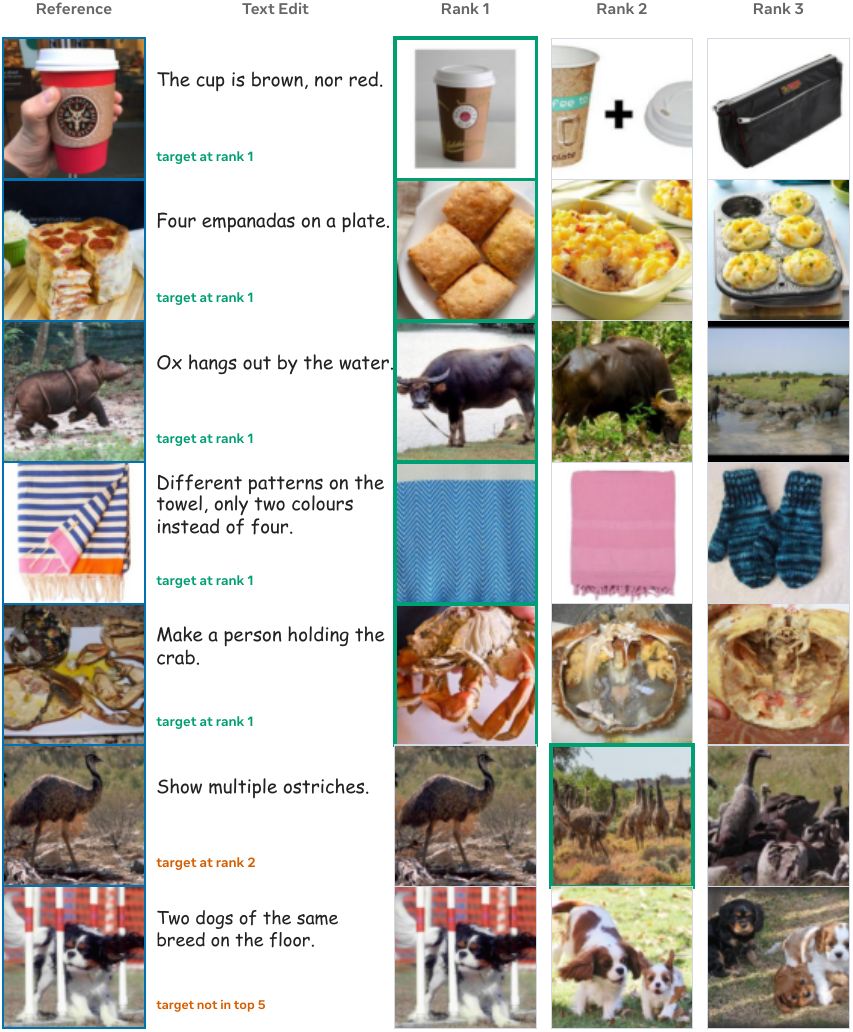}
\caption{\textbf{Composed queries and the top-3 retrieval results}.
Green borders indicate the ground-truth targets. While the overall retrieval results reflect that editing instructions correctly, the final two rows also highlight a zero-shot limitation: simultaneously preserving the reference category while applying count edits remains challenging. }
\label{fig:res:cirr}
\end{figure}

We evaluate latent space arithmetic on the CIRR dataset without training on composed queries, following the protocol outlined in Appendix~\ref{app:eval}. As shown in Table~\ref{tab:res:cirr}, using a prefix length of $K{=}1$ yields the strongest performance, demonstrating that an edit vector applied to a single token representation can effectively steer the global representation space. Figure~\ref{fig:res:cirr} shows the images retrieved with the raw image and editing instructions with FLAT zero-shot. 

\begin{center}
\centering
\captionof{table}{\textbf{Zero-shot CIRR composed-retrieval diagnostic across keep-lengths.} Each column uses the step size selected by Hit@1 from the common sweep in App.~\ref{app:eval}; Hit@5 and Hit@10 are evaluated at the same selected value.}
\label{tab:res:cirr}
\small
\setlength{\tabcolsep}{9pt}
\begin{tabular}{@{}lccccc@{}}
\cmidrule[\heavyrulewidth](l{-\tabcolsep}r{-\tabcolsep}){1-6}
Metric & $K{=}1$ & $K{=}4$ & $K{=}16$ & $K{=}64$ & $K{=}256$ \\
\cmidrule[\lightrulewidth](l{-\tabcolsep}r{-\tabcolsep}){1-6}
Hit@1  & \textbf{44.0} & 38.5 & 37.7 & 33.4 & 35.4 \\
Hit@5  & \textbf{75.2} & 70.1 & 70.5 & 69.2 & 71.5 \\
Hit@10 & \textbf{82.9} & 80.3 & 81.4 & 80.5 & 81.7 \\
\cmidrule[\lightrulewidth](l{-\tabcolsep}r{-\tabcolsep}){1-6}
\rowcolor{gray!12}
Selected $\alpha$ & 3 & 4 & 8 & 8 & 3 \\
\cmidrule[\heavyrulewidth](l{-\tabcolsep}r{-\tabcolsep}){1-6}
\end{tabular}
\end{center}

\clearpage

\section{Ablation on the training loss contributions}
\label{app:loss-ablation}
\begin{figure}[t]
\centering
\includegraphics[width=\textwidth]{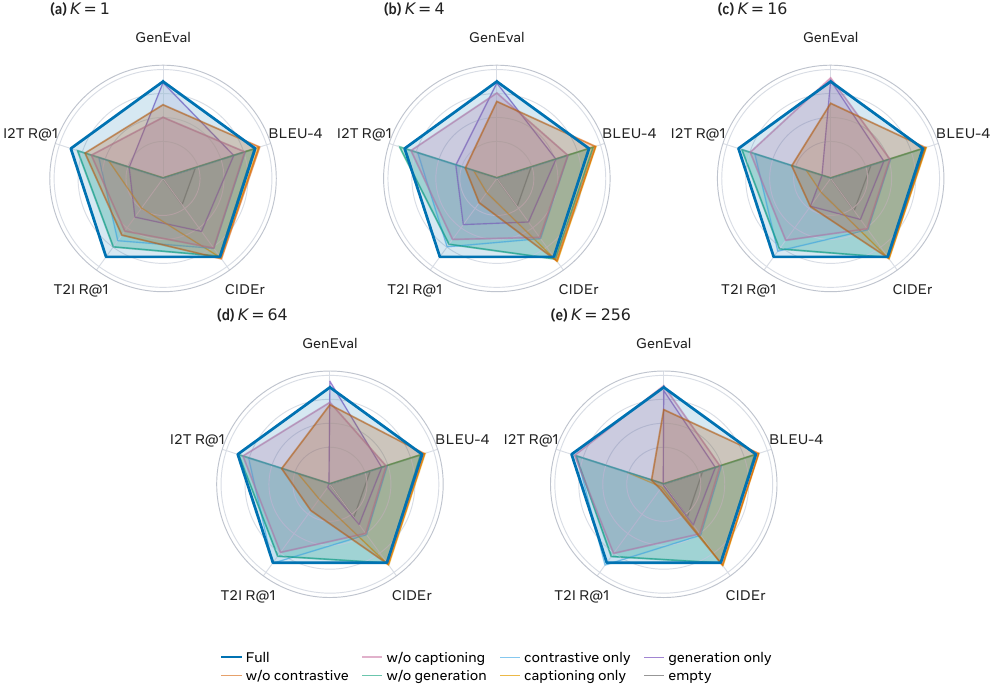}
\caption{\textbf{Ablation on training loss. Eval performance at
$K\in\{1,4,16,64,256\}$.} Each axis is normalized at the same prefix length for plot.}
\label{fig:app:loss-radar}
\end{figure}

FLAT pre-training incorporates three loss terms targeting retrieval, T2I generation, and I2T generation. We perform an ablation study by training models on different subsets of these losses to evaluate their individual and joint contributions. Table~\ref{tab:app:loss-ablation-allk} reports downstream scores across all evaluated prefix lengths and loss combinations. Each variant uses the same initialization and $40\mathrm{k}$-step schedule, varying only the active loss components. Incorporating all three losses provides a clear advantage, yielding balanced, optimal performance across all metrics and prefix lengths $K$. Figure~\ref{fig:app:loss-radar} plots the normalized evaluation scores for each ablated variant across all prefix lengths. 

\begin{table}[!htbp]
\centering
\caption{\textbf{Training loss ablation using all prefix lengths.}
$\mathcal{L}_{\mathrm{ctr}}$, $\mathcal{L}_{\mathrm{cap}}$, and
$\mathcal{L}_{\mathrm{gen}}$ denote contrastive, captioning, and image-generation
loss respectively. }
\label{tab:app:loss-ablation-allk}
\scriptsize
\setlength{\tabcolsep}{2.1pt}
\begin{minipage}[t]{0.49\textwidth}
\centering
\textbf{$K{=}1$}\\[1pt]
\begin{tabular}{@{}lrrrrr@{}}
\toprule
Objectives & I2T & T2I & B@4 & CIDEr & Gen. \\
\midrule
$\emptyset$ & 0.1 & 0.2 & 24.4 & 80.8 & 0.000 \\
$\mathcal{L}_{\mathrm{ctr}}$ & 50.7 & 57.2 & 33.9 & 117.3 & 0.001 \\
$\mathcal{L}_{\mathrm{cap}}$ & 48.2 & 46.3 & 35.9 & 123.3 & 0.004 \\
$\mathcal{L}_{\mathrm{gen}}$ & 41.7 & 49.6 & 32.1 & 106.6 & 0.328 \\
$\mathcal{L}_{\mathrm{ctr}}+\mathcal{L}_{\mathrm{cap}}$ & 56.1 & 59.2 & 35.3 & 122.2 & 0.004 \\
$\mathcal{L}_{\mathrm{ctr}}+\mathcal{L}_{\mathrm{gen}}$ & 52.6 & 54.1 & 33.8 & 117.2 & 0.280 \\
$\mathcal{L}_{\mathrm{cap}}+\mathcal{L}_{\mathrm{gen}}$ & 54.2 & 55.4 & 36.1 & 123.9 & 0.298 \\
All three & 57.8 & 62.4 & 35.4 & 122.7 & 0.329 \\
\bottomrule
\end{tabular}
\end{minipage}\hfill
\begin{minipage}[t]{0.49\textwidth}
\centering
\textbf{$K{=}4$}\\[1pt]
\begin{tabular}{@{}lrrrrr@{}}
\toprule
Objectives & I2T & T2I & B@4 & CIDEr & Gen. \\
\midrule
$\emptyset$ & 0.1 & 0.0 & 27.3 & 91.5 & 0.000 \\
$\mathcal{L}_{\mathrm{ctr}}$ & 50.8 & 57.0 & 34.1 & 118.1 & 0.002 \\
$\mathcal{L}_{\mathrm{cap}}$ & 26.4 & 21.2 & 38.7 & 133.7 & 0.005 \\
$\mathcal{L}_{\mathrm{gen}}$ & 41.8 & 50.1 & 32.1 & 107.0 & 0.323 \\
$\mathcal{L}_{\mathrm{ctr}}+\mathcal{L}_{\mathrm{cap}}$ & 55.1 & 56.2 & 38.3 & 131.9 & 0.005 \\
$\mathcal{L}_{\mathrm{ctr}}+\mathcal{L}_{\mathrm{gen}}$ & 52.4 & 54.7 & 34.1 & 117.4 & 0.309 \\
$\mathcal{L}_{\mathrm{cap}}+\mathcal{L}_{\mathrm{gen}}$ & 36.2 & 36.6 & 38.6 & 133.1 & 0.298 \\
All three & 53.9 & 60.1 & 37.6 & 130.4 & 0.325 \\
\bottomrule
\end{tabular}
\end{minipage}

\vspace{6pt}
\begin{minipage}[t]{0.49\textwidth}
\centering
\textbf{$K{=}16$}\\[1pt]
\begin{tabular}{@{}lrrrrr@{}}
\toprule
Objectives & I2T & T2I & B@4 & CIDEr & Gen. \\
\midrule
$\emptyset$ & 0.2 & 0.0 & 30.6 & 104.3 & 0.000 \\
$\mathcal{L}_{\mathrm{ctr}}$ & 50.4 & 56.7 & 34.1 & 117.7 & 0.002 \\
$\mathcal{L}_{\mathrm{cap}}$ & 28.0 & 19.8 & 40.3 & 138.3 & 0.004 \\
$\mathcal{L}_{\mathrm{gen}}$ & 10.1 & 40.6 & 32.9 & 110.4 & 0.325 \\
$\mathcal{L}_{\mathrm{ctr}}+\mathcal{L}_{\mathrm{cap}}$ & 53.5 & 56.0 & 40.0 & 137.2 & 0.003 \\
$\mathcal{L}_{\mathrm{ctr}}+\mathcal{L}_{\mathrm{gen}}$ & 51.5 & 53.4 & 33.9 & 117.0 & 0.328 \\
$\mathcal{L}_{\mathrm{cap}}+\mathcal{L}_{\mathrm{gen}}$ & 41.6 & 41.0 & 40.1 & 137.8 & 0.294 \\
All three & 54.3 & 58.4 & 39.7 & 136.8 & 0.324 \\
\bottomrule
\end{tabular}
\end{minipage}\hfill
\begin{minipage}[t]{0.49\textwidth}
\centering
\textbf{$K{=}64$}\\[1pt]
\begin{tabular}{@{}lrrrrr@{}}
\toprule
Objectives & I2T & T2I & B@4 & CIDEr & Gen. \\
\midrule
$\emptyset$ & 0.2 & 0.0 & 31.0 & 105.3 & 0.000 \\
$\mathcal{L}_{\mathrm{ctr}}$ & 50.4 & 56.9 & 34.0 & 117.7 & 0.003 \\
$\mathcal{L}_{\mathrm{cap}}$ & 35.6 & 21.0 & 40.7 & 139.3 & 0.004 \\
$\mathcal{L}_{\mathrm{gen}}$ & 1.1 & 4.4 & 33.0 & 110.4 & 0.327 \\
$\mathcal{L}_{\mathrm{ctr}}+\mathcal{L}_{\mathrm{cap}}$ & 52.7 & 54.8 & 39.9 & 137.8 & 0.004 \\
$\mathcal{L}_{\mathrm{ctr}}+\mathcal{L}_{\mathrm{gen}}$ & 51.7 & 53.7 & 33.8 & 116.8 & 0.299 \\
$\mathcal{L}_{\mathrm{cap}}+\mathcal{L}_{\mathrm{gen}}$ & 42.7 & 37.3 & 40.5 & 138.7 & 0.296 \\
All three & 52.9 & 56.7 & 40.1 & 137.5 & 0.319 \\
\bottomrule
\end{tabular}
\end{minipage}

\vspace{6pt}
\begin{minipage}[t]{0.49\textwidth}
\centering
\textbf{$K{=}256$}\\[1pt]
\begin{tabular}{@{}lrrrrr@{}}
\toprule
Objectives & I2T & T2I & B@4 & CIDEr & Gen. \\
\midrule
$\emptyset$ & 0.2 & 0.0 & 30.8 & 105.0 & 0.000 \\
$\mathcal{L}_{\mathrm{ctr}}$ & 51.0 & 57.0 & 34.2 & 118.0 & 0.003 \\
$\mathcal{L}_{\mathrm{cap}}$ & 39.4 & 4.0 & 40.8 & 139.7 & 0.004 \\
$\mathcal{L}_{\mathrm{gen}}$ & 0.6 & 1.1 & 33.1 & 110.8 & 0.323 \\
$\mathcal{L}_{\mathrm{ctr}}+\mathcal{L}_{\mathrm{cap}}$ & 51.2 & 54.5 & 40.0 & 137.8 & 0.003 \\
$\mathcal{L}_{\mathrm{ctr}}+\mathcal{L}_{\mathrm{gen}}$ & 51.5 & 53.6 & 33.9 & 117.1 & 0.329 \\
$\mathcal{L}_{\mathrm{cap}}+\mathcal{L}_{\mathrm{gen}}$ & 13.6 & 6.7 & 40.7 & 139.2 & 0.297 \\
All three & 52.2 & 56.3 & 40.2 & 137.5 & 0.327 \\
\bottomrule
\end{tabular}
\end{minipage}
\end{table}

\begin{figure}[!htbp]
\centering
\includegraphics[width=\textwidth]{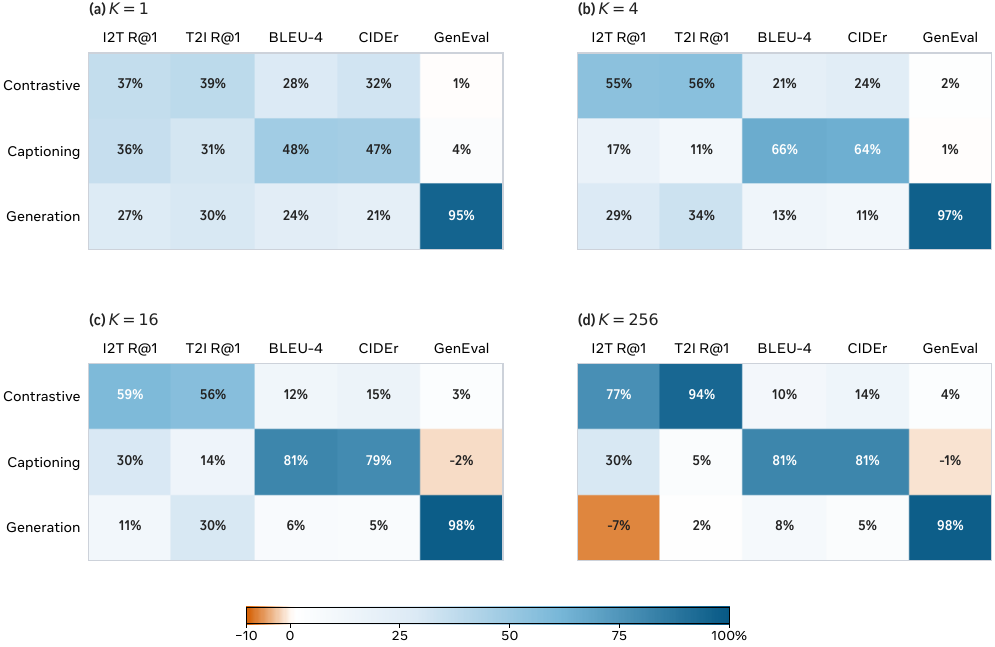}
\caption{\textbf{Relative Shapley attribution at different prefix lengths.}
Cell values are percentages of the total improvement over the untrained
checkpoint for each metric. Blue denotes positive attribution and orange denotes
negative attribution.}
\label{fig:app:loss-shapley}
\end{figure}

We also compute Shapley values~\citep{shapley1953value} to quantify the marginal contribution of each loss objective. For a loss component $i$ and downstream metric $v$, the Shapley attribution is defined as:
\begin{equation}
\phi_i(v)=\sum_{S\subseteq\mathcal{O}\setminus{i}}
\frac{|S|!(|\mathcal{O}|-|S|-1)!}{|\mathcal{O}|!}
\left[v(S\cup{i})-v(S)\right],
\end{equation}
where $\mathcal{O}$ is the set of all three losses.  Figure~\ref{fig:app:loss-shapley} visualizes these Shapley values across prefix lengths. The heatmaps normalize each attribution by $v(\mathcal{O})-v(\emptyset)$, ensuring that each metric column sums to $100\%$. The task-aligned attribution structure becomes sharper as the prefix widens: contrastive learning increasingly accounts for retrieval performance, captioning dominates both text-generation metrics from $K{=}16$ onward, and image generation consistently accounts for at least $95\%$ of GenEval. Minor negative entries at $K{=}16$ and $K{=}256$ indicate that auxiliary loss contributions are not uniformly positive across all prefix lengths, although the primary task-aligned attributions remain dominant.

\FloatBarrier

\section{Truncated prefixes and null padding}
\label{app:padding}

\begin{figure}[!htbp]
\centering
\includegraphics[width=\textwidth]{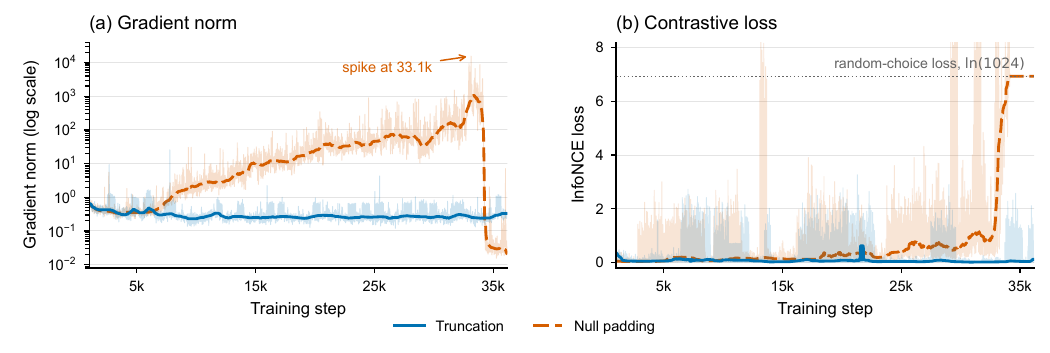}
\caption{\textbf{Optimization behavior under truncation and null padding.}
Curves are smoothed with a
$1{,}010$-step rolling median; faint lines show the unsmoothed measurements.
The null-padding run develops large gradient excursions before its contrastive
loss approaches the random-choice value.}
\label{fig:app:drop-mask-curves}
\end{figure}

\begin{figure}[!htbp]
\centering
\includegraphics[width=\textwidth]{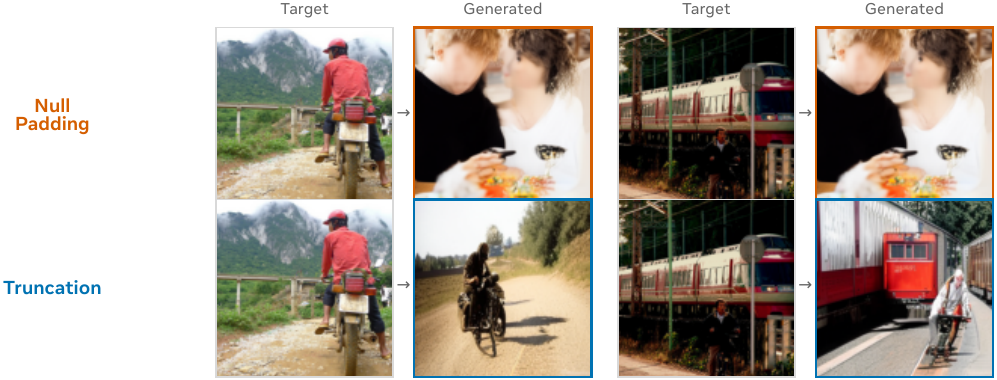}
\caption{\textbf{Images generated by prefix truncation v.s. null padding at step $40$k.}
Each column uses the same COCO target--caption pair and diffusion noise.}
\label{fig:app:drop-mask-samples}
\end{figure}

Section~\ref{sec:method:overview} describes truncating the register sequence to its first $K$ positions. An alternative strategy retains a fixed sequence length $N$ and replaces positions $i \ge K$ with learned null embeddings, as in FlexTok~\citep{bachmann2025flextok}. Empirically, we find that null-padded token sequences introduce optimization instabilities during FLAT joint training. Figure~\ref{fig:app:drop-mask-curves} illustrates these training dynamics. Under null padding, the gradient norm grows rapidly and spikes near $33\mathrm{k}$ steps, causing the contrastive loss to collapse toward $\ln(1024)$—the theoretical random-baseline value for the global batch. In contrast, prefix truncation remains stable throughout the same training window. As shown in Figure~\ref{fig:app:drop-mask-samples}, null-padded conditioning loses semantic correspondence between the source image and its generated counterpart, whereas prefix truncation preserves recognizable semantic relationships. Because these runs follow their respective original training recipes (including differing $K$-sampling granularities and learning-rate schedules), we interpret these findings as qualitative evidence of training dynamics.

\end{document}